\documentclass[acmsmall]{acmart}
\AtBeginDocument{%
  }

\usepackage[utf8]{inputenc} 
\usepackage[T1]{fontenc}    
\usepackage{hyperref}       
\usepackage{url}            
\usepackage{amsfonts}       
\usepackage{nicefrac}       
\usepackage{microtype}      
\usepackage{graphicx}
\usepackage{natbib}
\usepackage{subcaption}
\usepackage{algorithm}
\usepackage{algorithmic}

\usepackage{tikz}
\usepackage{xcolor}
\usepackage{pgfmath}
\usepackage{amsmath}
\usepackage{pgfplots}
\usepackage{pdfpages}
\usepackage{tikz}
\usepackage[absolute,overlay]{textpos}
\usepackage{blindtext}
\usepackage{relsize}
\usepackage{hyperref}

\begin{document}

\begin{textblock*}{21cm}(0cm,28cm)
    \begin{tikzpicture}[remember picture, overlay]
      \node (center) {};
      \path (center)+(8,26.5) node (name) {\small{\copyright 2025 Copyright held by the owner/author(s). This is the author's version of the work.}};
      \path (center)+(8,26.1) node (name) {\small{It is posted here for your personal use. Not for redistribution. The definitive version of record}};
      \path (center)+(8,25.7) node (name) {\small{was published in ACM Transactions on Embedded Computing Systems \href{https://doi.org/10.1145/3748722}{10.1145/3748722}.}};
    \end{tikzpicture}
  \end{textblock*}
    \begin{textblock*}{21cm}(1.20cm,2.60cm)
    \begin{tikzpicture}[remember picture, overlay]
      \draw (0,0) rectangle (13.5,1.5); 
    \end{tikzpicture}
  \end{textblock*}

\title{Optimization of DNN-based HSI Segmentation FPGA-based SoC for ADS: A Practical Approach}

\author{Jon Gutiérrez-Zaballa}
\email{j.gutierrez@ehu.eus}
\orcid{0000-0002-6633-4148}
\author{Koldo Basterretxea}
\email{koldo.basterretxea@ehu.eus}
\orcid{0000-0002-5934-4735}
\affiliation{%
\department{Department of Electronics Technology}
  \institution{University of the Basque Country (UPV/EHU)}
  \city{Bilbao}
  \country{Spain}
}

\author{Javier Echanobe}
\email{franciscojavier.echanove@ehu.eus}
\orcid{0000-0002-1064-2555}
\affiliation{%
  \department{Department of Electricity and Electronics}
  \institution{University of the Basque Country (UPV/EHU)}
  \city{Leioa}
  \country{Spain}
}

\renewcommand{\shortauthors}{Gutiérrez-Zaballa et al.}


\begin{abstract}
The use of hyperspectral imaging (HSI) for autonomous navigation is a promising field of research that aims to improve the accuracy and robustness of detection, tracking, and scene understanding systems based on vision sensors.
The combination of advanced computer algorithms, such as deep neural networks (DNNs), and small-size snapshot HSI cameras allows to strengthen the reliability of those vision systems.
Using HSI, some intrinsic limitations of greyscale and RGB imaging in depicting physical properties of targets related to the spectral reflectance of materials (metamerism) are overcome.
Despite the promising results of many published HSI-based computer vision developments, the strict requirements of safety-critical applications such as autonomous driving systems (ADS) regarding latency, resource consumption, and security are prompting the migration of machine learning (ML)-based solutions to edge platforms.
This involves a thorough software/hardware co-design scheme to distribute and optimize the tasks efficiently among the limited resources of computing platforms.
With respect to inference, the over-parameterized nature of DNNs poses significant computational challenges for real-time on-the-edge deployment.
In addition, the intensive data preprocessing required by HSI, which is frequently overlooked, must be carefully managed in terms of memory arrangement and inter-task communication to enable an efficient integrated pipeline design on a system on chip (SoC).
This work presents a set of optimization techniques for the practical co-design of a DNN-based HSI segmentation processor deployed on a field programmable gate array (FPGA)-based SoC targeted at ADS, including key optimizations such as functional software/hardware task distribution, hardware-aware preprocessing, ML model compression, and a complete pipelined deployment.
Applied compression techniques significantly reduce the complexity of the designed DNN to 24.34\% of the original operations and to 1.02\% of the original number of parameters, achieving a 2.86x speed-up in the inference task without noticeable degradation of the segmentation accuracy.
\end{abstract}



\maketitle

\section{Introduction}\label{sec:intro}
The application of deep learning techniques, especially fully convolutional networks (FCN) \cite{long2015fully}, has boosted the advances in the field of image segmentation \cite{mo2022review}, improving the ability of AI algorithms to accurately recognise objects within images across various application domains including medical imaging \cite{cui2022deep}, remote sensing \cite{terentev2022current} and food industry \cite{ozdougan2021rapid}, among others.
Nevertheless, a remarkable challenge arises when objects with different spectral signatures appear similar under specific lighting conditions, complicating object segmentation.
This phenomenon, known as metamerism, is a concern for many RGB-based image processing applications \cite{foster2006frequency}.

To address this phenomenon, recent studies have explored the use of hyperspectral imaging (HSI) as a robust and efficient potential solution to acquire spectral information across a wider range of wavelengths, providing the discriminative AI-based algorithm with richer input.
When discussing HSI, it is important to distinguish between two concepts that are often interchangeably used in the literature.
On the one hand, some researchers are evaluating whether HSI in the visible range yields better results than traditional RGB images.
On the other hand, there is a line of research investigating whether utilizing information beyond the visible spectrum, regardless of whether it involves HSI, can lead to more accurate and robust segmentations.
Regarding the potential benefits of using HSI over RGB, the authors of \cite{gutierrez2023chip} review several studies from different industries, such as agriculture, food assessment, healthcare, and automotive, where superior performance is achieved by using HSI.
More recently, \cite{hanson2023hyper} for robot navigation and \cite{ren2024point} for facade segmentation have also found that HSI leads to clearer decision boundaries between classes.

Currently, the potential for HSI to be applied to dynamic environments such as autonomous driving systems (ADS), where accurate and timely data interpretation is crucial for ensuring passengers’ safety, has become more feasible due to the emergence of compact snapshot hyperspectral cameras \cite{10.1117/12.2077583, 9483975, 10.1117/12.2037607}.
This technology allows for the simultaneous capture of object reflectance across a multitude of wavelengths in a single shot at video rates.
Nonetheless, hyperspectral sensors, and especially snapshot sensors, usually require a computationally costly preprocessing stage to convert the acquired 2D raw data into 3D hyperspectral cubes.
This step is often overlooked by neural network developers, as the images used for training and testing are typically preprocessed beforehand.
In addition to this, the promising results of combining HSI with deep neural networks (DNNs) very often come at the cost of using over-parameterized deep learning models, leading to high computational complexity.
These DNNs typically contain millions of parameters and require the execution of billions of computation operations (floating-point, FLOPS or integer, OPS) per inference, posing significant challenges for the on-the-edge deployment of safety-critical applications.

In order to efficiently implement a complete segmentation pipeline in an embedded computing platform, careful planning using a refined hardware/software co-design methodology is required.
This scheme is essential to optimize the efficient integration of the different stages of the complete processing pipeline: raw data to hyperspectral cube preprocessing, data storage and arrangement in memory, data communication, and DNN inference.
Besides, a key consideration in this process is the identification and mitigation of potential bottlenecks that may limit overall performance.
Therefore, it is imperative to develop optimized solutions that can ensure reliable and fast performance.

Given these challenges, implementing these models on field programmable gate arrays (FPGAs) and FPGA-based system on chips (SoCs) emerges as a promising solution.
FPGAs and programmable SoCs enable the design of domain-specific processors tailored to each use case, facilitating optimized resource usage, power efficiency, and low latency while allowing for reconfigurability when needed.
This adaptability makes these platforms ideal for deploying advanced HSI-based segmentation models in resource-constrained applications, paving the way for more effective and reliable systems in ADS applications and beyond.

\begin{figure}[h!]
\centering
\includegraphics[width=13.5cm]{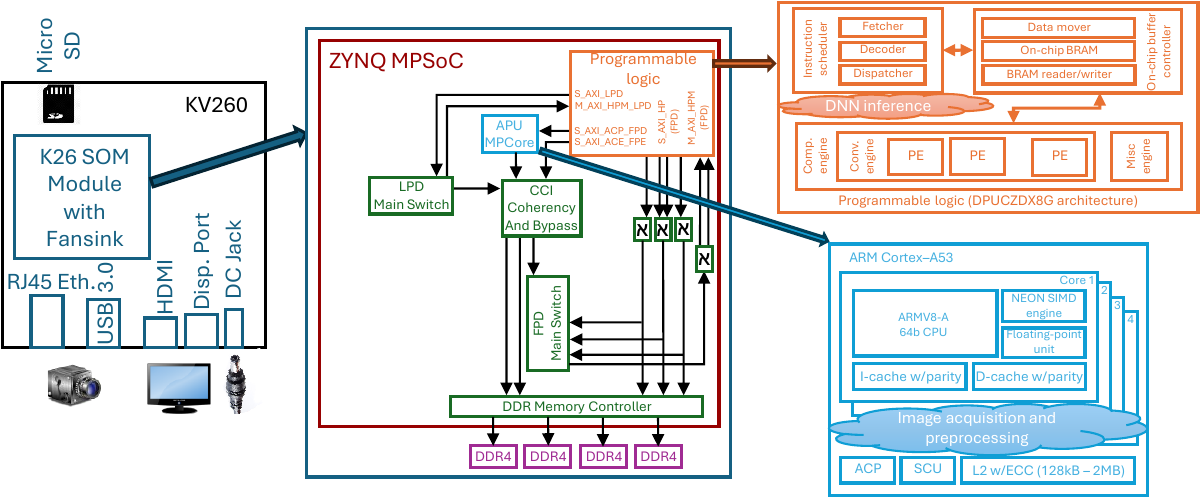}
\caption{Diagram of the DNN-based segmentation pipeline.
Left, KV260 board, adapted from \cite{k26somIdeal}; centre, Zynq UltraScale+ MPSoC, adapted from \cite{k26datasheet}, and right, Application Processing Unit, adapted from \cite{CortexA53} and Programmable Logic, adapted from \cite{dpu}.}
\label{fig:BlockDiagram}
\end{figure}

In this article, a holistic design approach for a DNN-based HSI segmentation pipeline optimized for FPGA-based SoCs is presented.
The target platform has been AMD-Xilinx’s KV260 board (see Figure \ref{fig:BlockDiagram}), which is tailored for edge vision applications and integrates the K26 SOM with a Zynq UltraScale+ MPSoC, where the segmentation performance and optimizations were evaluated.

This hardware/software co-design methodology and DNN architecture selection are aligned with the platform constraints to maximize efficiency.
The optimization techniques applied throughout the design process, from raw data preprocessing to DNN inference deployment are detailed.
Preprocessing stages leverage data- and thread-level parallelism, with some steps offloaded to hardware when feasible.
Computational profiling identifies suitable DNN compression techniques and assesses the need for multi-stage pipelined preprocessing to mitigate bottlenecks.

For DNN inference, the rigidity of the K26 SOM’s accelerator in quantized parameter representation and its lack of support for sparse matrix multipliers led to focus on channel pruning.
The proposed iterative pruning method combines static and dynamic analyses to define pruning targets and assess their feasibility, while ensuring minimal impact on inference quality.
It is also explained how to assess whether the initially set pruning ratio is excessive, or if further pruning can be applied in subsequent iterations without degrading performance.

Applied to this segmentation U-Net for the HSI-Drive v2.0 dataset \cite{gutierrez2023hsi, zenodov2_0}, this optimization scheme reduces inference operations by an order of magnitude and the number of parameters by two orders of magnitude, while preserving performance.
It also improves resource and power efficiency, making deployment more practical.

The rest of the article is organized as follows.
Section \ref{sec:relatedWork} covers the related work, including current HSI databases for ADS, state-of-the-Art (SotA) deep learning models for ADS and current pruning-based DNN model optimization strategies.
In Section \ref{sec:modelDevelopment}, the training and testing dataset, HSI-Drive v2.0 \cite{gutierrez2023hsi, zenodov2_0}, is described as well as the modified version of the original U-Net \cite{ronneberger2015u}.
The testing results against a SotA model are also compared there.
Section \ref{sec:modelOptimization} details the compressing operations applied to the baseline model, with particular emphasis on both the static and dynamic analyses of the model and the iterative pruning methodology.
The preprocessing of the raw images is explained in Section \ref{sec:rawImagePreprocessing}, which also includes a discussion about the optimal memory arrangement of the hyperspectral cubes.
Finally, Section \ref{sec:DeploymentAndTesting} provides details on the deployment of the complete pipeline (including raw data loading, preprocessing, cube transmission, and segmentation) on the KV260 board.
Different configurations of the pipeline with 1, 2 and 3 stages are presented and the performance of varying configurations of the K26's deep processing unit (DPU) is characterized in terms of latency, throughput and power consumption.
The article concludes with a discussion of the key findings in Section \ref{sec:conclusions}.

\section{Related Work}\label{sec:relatedWork}
The development of DNN-based HSI segmentation systems that are suitable for deployment on ADS embedded platforms requires firstly, the availability of a training dataset specifically designed for this task.
The performance of the resulting model, which is preferably based on network architectures that will eventually do not demand excessive computing, must be compared against SotA segmentation models on reference benchmarks available to the scientific community.
Finally, the obtained baseline models must be optimized for hardware implementation, a process that must be guided by the specifications of the target application.
In this section, a brief review of previous work published in this regard is provided.

\subsection{Hyperspectral Datasets for Autonomous Driving Systems}
\textbf{Hyko and Hyko v2.0.} 
The authors have acquired both VIS and NIR images using two snapshot mosaic imaging cameras \cite{winkens2017hyko}.
However, most NIR images, primarily of asphalt, are usually omitted for the experiments \cite{theisen2024hs3}.
The latest version contains 371 VIS images (254x510x15) that depict both urban and rural scenes and are labelled to distinguish between 10 different classes.
More than half (58.3\%) of the papers that cite Hyko mention it as part of the related work, typically referencing it as an example of other applications using HSI or as another dataset focused on HSI segmentation in ADS.
In addition to this, about 16.7\% of the papers have used Hyko to test the effectiveness of their demosaicing methods for snapshot HSI cameras.
Finally, the remaining 25\% of the works employ the dataset for neural network training and testing, but they do not include any discussion regarding model optimization or deployment on hardware.

\textbf{Hyperspectral City and Hyperspectral City v2.0.} 
The authors collected 1,330 images covering the 450-950nm VIS-NIR spectral range using LightGene camera sensor and contain up to 19 classes from urban scenarios \cite{you2019hyperspectral, huang2021hsicityv2}.
The images have high spatial (1889x1422) and spectral (128) resolution, so each hyperspectral cube takes more than 1GB, making on-board processing difficult.
More than half (62.5\%) of the papers that reference Hyperspectral City do so in their related work sections, while the remaining 37.5\% of the papers use the dataset for training or testing their neural networks, but none report its deployment on hardware.

\textbf{HSI Road.} 
This dataset contains both 192x384 RGB and 25-band VNIR (680-960nm) images captured with a progressive scan colour camera and a snapshot mosaic imaging camera respectively \cite{lu2020hsi}.
Even though the 3,799 images are from rural and urban scenes, the authors only differentiate between two classes: background and road.
Nearly all (92.86\%) papers cite HSI Road in their related work or introduction sections.
In addition, it has been cited once in a review on sensing technologies and once in a paper where it was used as a training/testing database, though no hardware deployment has been reported.

\textbf{HSI-Drive and HSI-Drive v2.0.} 
HSI-Drive contains 752 cubes (216x409x25) captured with a snapshot mosaic imaging camera covering the 535-975nm spectral range \cite{basterretxea2021hsi, gutierrez2023hsi, zenodov2_0}.
The images contain 10 categories and are structured according to road type, time of day, season, and environmental conditions.
Apart from previous studies, HSI-Drive has mainly been cited in reviews discussing sensing for ADS in variable weather conditions (20\%) or as part of the related work, highlighting other HSI application areas or datasets focused on HSI segmentation in ADS (80\%).
Similarly, all citations of HSI-Drive v2.0 fall into this latter category.
No prior work has focused on the training or efficient deployment of HSI-based semantic segmentation processors.

\textbf{HyperDrive.} 
Although HyperDrive is primarily designed for robots navigating in unstructured environments, its relevance to this application warrants mention \cite{hanson2023hyper}.
It includes data captured by two VNIR and NIR point spectrometers, along with two snapshot mosaic VNIR and SWIR imaging cameras, producing hyperspectral cubes of 1012x1666 pixels with 33 spectral bands, covering the 660-900nm and 1100-1700nm ranges respectively.
The dataset is complemented by high-resolution RGB images acquired with an Allied Vision camera.
Since this database is relatively new, few studies have utilized it: just one paper references it in the related work section, while another uses it for model training, although no hardware deployment has been reported.

Overall, although some articles have used these datasets to train neural networks for ADS scene segmentation, none report implementations on embedded computing platforms.

\subsection{State-of-the-Art Deep Neural Networks for computer vision-based Autonomous Driving Systems}\label{sec:sotaModels}
Prior to this research, the only HSI benchmark for ADS applications was based on the Hyperspectral City v2.0 dataset \cite{huang2021hsicityv2}, with few competing models.
A second benchmark, HS3-Bench \cite{theisen2024hs3}, was proposed afterward, but it is too recent to be considered a reference.
This new benchmark combines three HSI datasets that are radically different in nature: Hyko, Hyperspectral City, and HSI-Drive, all in their most up-to-date versions.
Consequently, it appears challenging to find a model that achieves optimal results across all three datasets.
However, when analysing just the HSI-Drive dataset part, the best results are obtained with a regularized U-Net, a DNN which is very similar to the one employed in this study.
No details are provided regarding any implementation on embedded hardware.

Since there is no reference HSI ADS benchmark, several RGB-based benchmarks with similar applications were reviewed (e.g. CityScapes \cite{Cordts2016Cityscapes}, BDD100K \cite{yu2020bdd100k}, CamVid \cite{brostow2009semantic}, Mapillary Vistas \cite{neuhold2017mapillary}, and Apolloscape \cite{huang2018apolloscape}).
Given that CityScapes is the most cited dataset on Google Scholar, it was selected as the reference for model comparison.
At the time of writing, Intern Image \cite{wang2023internimage} was the highest-ranked model among those sharing code for reproducibility \cite{cityscapes_benchmarks}, making it the choice.
Intern Image \cite{wang2023internimage} is based on a deformable convolution operation that allows for a big receptive field which may be necessary for image detection and segmentation tasks where it could be beneficial to aggregate long-range spatial information.
The basic block composes layer normalization, feed forward network and Gaussian error linear unit activation layer.
The rest of the SotA models that share some details about their architecture are all based on vision language models as the case of VLT \cite{hummer2023vltseg} (third position in both rankings) but, as the authors admit, those models are not applicable to real-time settings \cite{hummer2023vltseg}.

\subsection{Current Pruning Methods}\label{sec:current_pruning_methods}
Model pruning is based on the idea of removing the least significant parameters from a model based on specific criteria.
Currently, the scientific community agrees that pruning a large, sparse model generally yields better results than training a smaller, dense model from scratch \cite{zhu2017prune}.
As a result, various pruning methods have emerged which can be differentiated by three main characteristics: sparsity, criteria, and time.

Pruning sparsity can be classified as fine-grained/sparse/unstructured, when individual weights are set to zero, or as coarse-grained/dense/structured, when entire filters or layers are removed from the computational graph.
Fine-grained pruning typically allows for more weights to be removed without significantly harming performance, but is only effective if the underlying hardware can optimize operations with sparse matrices.
Regarding the criteria, there are several options that are commonly used nowadays, such as L1, L2, cosine similarity \cite{shao2021dynamic}, or change in loss \cite{molchanov2019importance}.

Pruning has traditionally been performed after model training for image classification, followed by fine-tuning to mitigate accuracy loss.
Given its effectiveness, this technique continues to be employed and refined today, as demonstrated in \cite{NEURIPS2022_1caf09c9}, which explores accurate post-training pruning.
Similarly, pruning has been adapted for modern transformer-based models, as shown in \cite{NEURIPS2022_987bed99} and \cite{zhang2024plug}.

The lottery ticket hypothesis \cite{frankle2018lottery} sparked interest in pre-training and training-aware pruning to optimize models more efficiently, particularly for classification tasks.
It proposes that within a randomly initialized dense network, a subnetwork exists that, when trained in isolation, matches the original model’s performance.
This idea was challenged by \cite{liu2018rethinking}, who showed that re-randomized subnetworks could perform even better.
Subsequently, \cite{frankle2018lottery} introduced iterative pruning, achieving superior results over one-shot pruning and complicating the notion of early winning ticket identification.
The hypothesis was revisited in \cite{zhou2019deconstructing}, suggesting that randomly initialized subnetworks could perform well without further training, shifting focus to efficient subnetwork search.
Building on this, \cite{ramanujan2020s} improved the search process, identifying untrained subnetworks nearing state-of-the-art performance on datasets like CIFAR-10 and ImageNet.
Finally, \cite{malach2020proving} formalized the strong lottery ticket hypothesis under overparameterization and weight distribution assumptions, though highlighting the computational difficulty of finding such subnetworks.

Despite its progress, network pruning faces several challenges.
Most studies focus on image classification with small datasets, limiting generalizability.
This raises the question of whether tasks like semantic segmentation, explored in this article, could similarly benefit, a topic still underexplored \cite{zhu2025comprehensive}.
Moreover, the lack of unified benchmarks hampers fair comparison across methods.
To address this, the authors of \cite{cheng2024survey} maintain a repository compiling pruning-related papers and open-source implementations.

Given the scope of this work, the pruning approaches tailored to segmentation tasks are highlighted, particularly those applied before or during training, which remain less explored.
For a comprehensive review of general pruning methods, readers are referred to \cite{zhu2025comprehensive} and \cite{cheng2024survey}.

Pruning at initialization for image segmentation has recently been explored in \cite{ditschuneit2022auto} and \cite{iurada2024finding}.
The authors of \cite{ditschuneit2022auto} found that pruning at initialization led to slightly inferior performance due to reduced model capacity, with the effect expected to be more pronounced on complex datasets.
Similarly, \cite{iurada2024finding} investigated pruning at initialization for pre-trained models, concluding that dense network performance could only be approximated at very low sparsity levels.

Pruning during training is closely tied to data variability, which affects training stability and convergence.
For example, the authors of STAMP \cite{dinsdale2022stamp} applied pruning during training on the Medical Decathlon datasets, benefiting from the low variability in MRI and CT scans, leading to smoother convergence.
The authors of \cite{ditschuneit2022auto} also showed that pruning at the end of training causes a significant performance drop, although the absence of a reported stopping criterion raises the possibility of overfitting contributing to this decline.
Finally, the success of transformer-based segmentation models has driven the development of specialized pruning strategies, as shown by \cite{bai2022dynamically} and \cite{yang2023pruning}.
However, these methods still result in models with millions of parameters and require substantial computational resources, often more than a GPU-day, to search for pruned architectures.

\section{Model's Development}\label{sec:modelDevelopment}
For this study, a U-Net architecture was selected as the reference DNN for segmentation.
The encoder-decoder network was trained and evaluated on the HSI-Drive v2.0 dataset \cite{gutierrez2023hsi, zenodov2_0}.

\subsection{Evaluation Dataset}
\begin{figure}[h!]
\centering
\begin{subfigure}{0.48\linewidth}
\centering
\includegraphics[width=5.75cm]{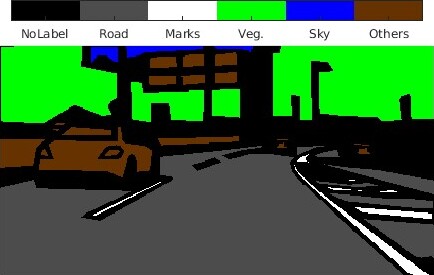}
\caption{Ground truth.}
\label{fig:gt}
\end{subfigure}
\begin{subfigure}{0.48\linewidth}
\centering
\includegraphics[width=5.75cm]{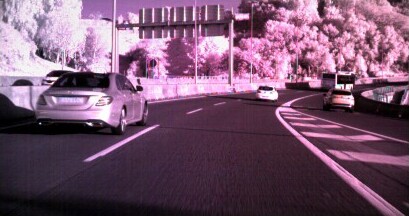}
\caption{False-RGB.}
\label{fig:falsergb}
\end{subfigure}
\caption{Ground truth (left) and false-RGB (right) of image 721, an example of HSI-Drive v2.0 \cite{gutierrez2023hsi, zenodov2_0}.}
\label{fig:exampleImage721}
\end{figure}

HSI-Drive v2.0 (Figure \ref{fig:falsergb}) is a hyperspectral imaging database acquired with a small-size snapshot HSI camera.
It contains more than 750 manually annotated images of real driving scenarios that aims to facilitate research on the use of HSI in the development of ADS.

In this experiment, the ground truth (Figure \ref{fig:gt}) is composed of 5 classes: Road (Tarmac), Marks (Road Marks), Vegetation, Sky and Others.
As Vegetation and Sky are usually part of the surroundings and the background, the experiment allows for the detection of potential obstacles such as vehicles, cyclists or pedestrians which may demand responsive actions.
This, in turn, could enhance ADS capabilities like emergency braking or collision alert systems.

\subsection{Model Architecture, Training, and Data Stratification}\label{sec:modelArchitectureTraining}
The base image segmentation DNN in this study is U-Net \cite{ronneberger2015u}, an encoder-decoder FCN which has been adapted to perform semantic segmentation on HSI.
Unlike the original U-Net, which was designed for two-class RGB biomedical image segmentation with a 4-level encoder-decoder and 64 initial filters, this adaptation optimizes the architecture for 25-band HSI.

The hyperparameters of the reference non-compressed model were optimized via a grid search on encoder-decoder depth and the initial number of filters, as detailed in \cite{gutierrez2023hsi}.
The resulting DNN consists of a 5-level encoder-decoder architecture, where each basic block comprises a 2D convolutional layer ($conv2D$) followed by a batch normalization (BN) (not present in the original U-Net) and ReLU activation.
Each encoder and decoder level, as well as the base, contains two of these blocks.
In the encoder, the basic block is preceded by a 2D max-pooling layer (except for the input layer) and ends with a dropout layer (also not present in the original U-Net).
In the decoder, the basic block is preceded by a transposed $conv2D$.
The first convolutional block contains 32 filters instead of 64, and the same strategy as in the original U-Net is maintained: doubling the number of filters while reducing the spatial resolution in the encoder and halving the number of filters while increasing the resolution in the decoder (Figure 2 in \cite{gutierrez2024evaluating}).
This U-Net thus contains 31.125 M parameters, occupying 118.73 MB of memory, and requires 34.90 GFLOPS per inference to segment each image frame.

The training, validation and testing of the models has been done following a stratified 5-fold cross-validation.
This implies dividing the dataset into 5 subsets (3 for training, 1 for validation and 1 for testing) and then performing 5 different training/validation/testing rounds where in each round the folds are shifted from subset to subset.

Training has been performed on a Dell Precision Tower 7920 featuring an Intel Xeon Silver 4216 CPU (16 cores, 32 threads and 2.10 GHz) equipped with an NVIDIA GeForce RTX 3090 with 24GB of GDDR6 VRAM.
The training, repeated 3 times to avoid biased results from poor random Glorot initialisation, has completed a maximum of 200 epochs (early stopping was activated) using Adam optimizer with an initial learning rate of 0.001 and employing shuffled mini-batches of size 30.
For more information regarding the model's architecture and the training procedure, the reader is referred to \cite{gutierrez2024evaluating}.

\subsection{Comparison Against SotA Models}\label{sec:sota}
The simplicity of the U-Net layers, compared to the complex, novel operators of some SotA models, makes it more appropriate for implementations on embedded computing devices.
However, it is worth considering whether the use of more sophisticated models could lead to higher performance in terms of accuracy.
To address this question, InterImage \cite{wang2023internimage} was selected, a SotA segmentation model that ranks highest among those with publicly available code for reproducibility in the CityScapes benchmark \cite{cityscapes_benchmarks}.
Selecting T variant and modifying its input and first convolutional block to make it compatible with HSI results in a model with 58.97 M parameters, a size of 224.95 MB and 66.05 GFLOPS, which almost doubles the figures of U-Net.

Regarding segmentation metrics, the two models are compared in Table \ref{tab:FloatingUnet_InternImage} in terms of Intersection over Union (IoU), since it provides a balanced measure that accounts for both precision and recall.
As global metrics are biased towards the most predominant class, a result of real-world data distribution, class-weighted metrics are also calculated mitigating the over-representation of large classes.
As for U-Net, the obtained precision is higher than 90\% for every class and both the recall and the IoU are higher than 85\% except for the most diverse class, Other.
Comparing both models, this U-Net model outperforms the SotA model in terms of global IoU (gIoU) and weighted IoU (wIoU), that is, the performance ceiling is constrained by the dataset quality and quantity rather than by the model's architectural complexity.
Since applying a stratified 5-fold cross-validation aims to verify the invariance of the model's robustness to different partitions of the dataset, from now on, and without loss of generalization, the results provided will correspond to the fifth fold.

\begin{table*}[!h]
\centering
\caption{IoU for the 32-bit floating-point U-Net model (left) and Intern Image \cite{wang2023internimage} model (right).}
\label{tab:FloatingUnet_InternImage}
\resizebox{13.5cm}{!}{%
\begin{tabular}{ccccccccccccc}
\toprule
\multicolumn{1}{c}{} & \multicolumn{6}{c}{\textbf{U-Net}} & \multicolumn{6}{c}{\textbf{Intern Image}} \\
\cmidrule(lr){2-7} \cmidrule(lr){8-13}
\textbf{Class / Fold} & \textbf{Fold 1} & \textbf{Fold 2} & \textbf{Fold 3} & \textbf{Fold 4} & \textbf{Fold 5} & \textbf{Mean} & \textbf{Fold 1} & \textbf{Fold 2} & \textbf{Fold 3} & \textbf{Fold 4} & \textbf{Fold 5} & \textbf{Mean} \\
\midrule
\textbf{Road}       						& 97.32           & 97.12           & 97.97           & 97.41           & 97.84           & \textbf{97.53}         & 97.54           & 96.80           & 97.70           & 97.57           & 97.97           &         97.51 \\
\textbf{Marks}      						& 82.13           & 83.13           & 89.01           & 87.15           & 88.30           & \textbf{85.94}         & 80.04           & 80.10           & 83.02           & 83.60           & 84.97           &         82.35 \\
\textbf{Vegetation} 						& 95.53           & 93.58           & 95.97           & 95.99           & 94.15           & \textbf{95.04}         & 95.60           & 92.08           & 94.87           & 96.10           & 95.34           &         94.80 \\
\textbf{Sky}      						    & 91.71           & 93.09           & 92.26           & 94.64           & 93.38           &         93.02          & 94.25           & 93.75           & 88.95           & 95.25           & 94.76           & \textbf{93.39} \\
\textbf{Other}      						& 78.16           & 77.63           & 81.82           & 77.94           & 77.38           & \textbf{78.59}         & 77.68           & 72.70           & 78.96           & 79.45           & 80.09           &         77.78 \\
\textbf{Global}     						& 94.43           & 93.73           & 95.48           & 94.90           & 94.67           & \textbf{94.64}         & 94.76           & 93.15           & 94.52           & 95.20           & 95.35           &         94.59 \\
\textbf{Weighted}   						& 85.28           & 85.87           & 89.41           & 88.65           & 88.40           & \textbf{87.52}         & 84.17           & 82.90           & 84.76           & 84.17           & 87.21           &         84.64 \\
\bottomrule
\end{tabular}}
\end{table*}

\section{Compression Techniques for DNNs in Co-design Environments}\label{sec:modelOptimization}
Although the developed U-Net model is lighter than the most sophisticated segmentation models, it remains too heavy for deployment on embedded platforms when low-latency processing is required.
Therefore, as is common practice, compression techniques must be applied to obtain an optimized model.
In this section, a brief overview of the quantization scheme employed is provided, and then the focus is placed on the presented pruning strategy.

\subsection{Post Training Quantization}\label{sec:quantization}
The quantization strategy employed to transform the model from high accuracy floating-point arithmetic to 8-bit integer operations is detailed in \cite{icecs2023}.
Briefly, a clipping of the input cube based on the data distribution of each of the 25 spectral channels is performed which allows to save 3 integer bits and augment the precision of the binary representation of the fixed-point values accordingly.
This ensures that 99.95\% of the data are correctly represented.

The customized quantization pipeline consists of the following stages: symmetric quantization for inputs, bias (both with Min-Max method) and weights (Min-MSE method) and asymmetric for activations (Min-MSE method too).
Additional applied techniques have been BN folding (to fuse a $conv2D$ and a BN layer and reduce the number of parameters) and cross layer equalization, minimizing the difference in the magnitude of the elements in the same layer or tensor without having to perform the costly per-channel quantization.
In addition to this, the quantization scheme has been homogeneous and uniform, the scale factor is restricted to be a power of two, and the quantization granularity is per-tensor.
The comparative figures after quantization are summed up in Table \ref{tab:postTrainingQuantization} and show that there has been almost no degradation in the IoU after the custom quantization of the U-Net.

\begin{table*}[!h]
\centering
\caption{IoU of the U-Net models according to the data representation.}
\label{tab:postTrainingQuantization}
\resizebox{13.5cm}{!}{%
\begin{tabular}{ccccccccccc}
\toprule
\textbf{Data representation} & \textbf{Mparams} & \textbf{GFLOPS} & \textbf{Size (MB)} & \textbf{Road} & \textbf{Marks} & \textbf{Vegetation} & \textbf{Sky} & \textbf{Other} & \textbf{Global} & \textbf{Weighted} \\
\midrule
\textbf{FP32}				 & 31.10            & 34.87			  & 118.73					 & 97.84		 & 88.30			   & 94.15				 & 93.38		& 77.38			 & 94.67		   & 88.40 \\
\textbf{INT8}				 & 31.10            & 34.87			  & 29.66					 & 97.65		 & 87.99               & 93.46               & 92.38		& 76.57		     & 94.27           & 87.82 \\
\bottomrule
\end{tabular}}
\end{table*}

After the quantization of the model, its memory requirements have considerably decreased to a fourth of the initial one, but the number of GFLOPS to be performed per-inference has not been reduced.
However, reducing the bit-width permits vectorization as INT8 MAC operations could be performed in parallel in some processors, taking advantage of SIMD (Single Instruction, Multiple Data) computing strategy (vector registers, Neon (ARM) \cite{ARM2020Neon} or DSP48E2 slices (AMD-Xilinx) \cite{fu2016deep}).
In this regard, the acceleration is tightly coupled with how the data are read/stored from/to memory as this is usually the most time consuming operation.
Therefore, in order to trim the number of FLOPS, the use of pruning methods has been investigated.

\subsection{Iterative Structured Pruning}
In a typical convolutional neural network for object detection, pruning a specific convolutional filter reduces the size of the output feature map, so only the weights and BN parameters corresponding to the pruned feature dimension need to be removed.
However, modern DNNs for image segmentation, such as U-Net, include concatenation blocks that merge features extracted at different kernel sizes and depths.
These concatenation layers, along with skip-connections linking encoder and decoder blocks, create dependencies between layers that make the pruning process more complex.
Because of these skip-connections, pruning in U-Net requires careful handling of concatenation layers to ensure the architecture remains consistent.
In contrast, other layers such as 2D max-pooling, dropout, and activation layers do not need explicit treatment.

In the following, the proposed post-training iterative pruning strategy tailored for U-Net segmentation models that comprises both static and dynamic analyses is described.
First, the static analysis focuses on the distribution of computational load and memory footprint across the DNN.
The pruning objective is established by identifying the most computationally intensive operations and considering the application requirements.
Subsequently, the dynamic analysis examines the model’s sensitivity to various pruning ratios (\textit{pr}) of the chosen objective, allowing the identification of layers that are most robust or over-parameterized.
These two analyses form the foundation of the proposed iterative pruning method, which begins with the selection of an overall \textit{pr}.
This ratio must be determined while taking into account the flexibility of each layer, as explained further below.

\subsubsection{Static Analysis}\label{sec:staticAnalysis}
The static analysis begins by evaluating the contribution of every layer in terms of FLOPS and number of parameters.
In the reference U-Net architecture, the operations demanding the highest FLOPS are the $conv2D$ and the transposed $conv2D$ layers.
In fact, other operations such as bias addition or BN account for only about 0.21\% of the total computational load.
Therefore, an accurate overall estimation can be obtained by focusing solely on these two convolution operations.
Equation \ref{equ:flopsTotal} represents the approximate number of FLOPS of the whole model

\begin{equation}
    FLOPS \approx \sum_{j=0}^{n} FLOPS_{conv_j} = \sum_{j=0}^{n} o_{h_{j}} \cdot o_{w_{j}} \cdot o_{f_{j}} \cdot k_{h_{j}} \cdot k_{w_{j}} \cdot i_{c_{j}} \cdot 2 \cdot (\frac{1}{4})_j
    \label{equ:flopsTotal}
\end{equation}

\noindent where ${}_j$ indicates the number of the $conv2D$ layer, $o_{h}$ is the height of the output feature map, $o_{w}$ is the width of the output feature map, $o_{f}$ is the number of filters or depth of the output feature map, $k_{h}$ is the height of the convolution kernel, $k_{w}$ is the width of the convolution kernel, $i_{c}$ is the number of input channels or depth of the $conv2D$ kernel and the factor 2 is necessary to account for the addition that have to be made during $conv2D$ operations (1 MAC = 2 FLOPS).
The $\frac{1}{4}$ factor is only applied to transposed $conv2D$ as they have a stride of 2 in both directions.

\begin{figure}[h!]
\centering
\begin{subfigure}{0.48\linewidth}
\centering
\includegraphics[width=6.25cm]{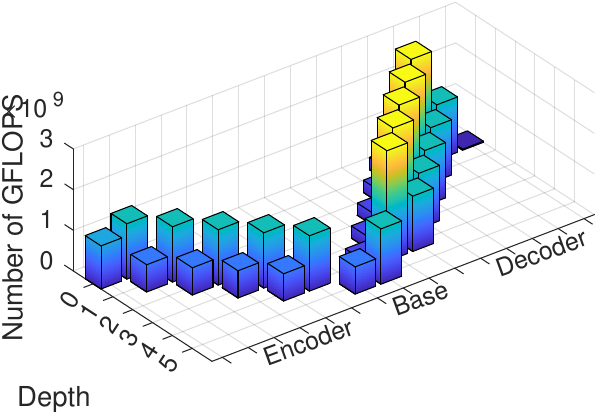}
\caption{Number of FLOPS.}
\label{fig:3dbarNumFLOPS}
\end{subfigure}
\begin{subfigure}{0.48\linewidth}
\centering
\includegraphics[width=6.25cm]{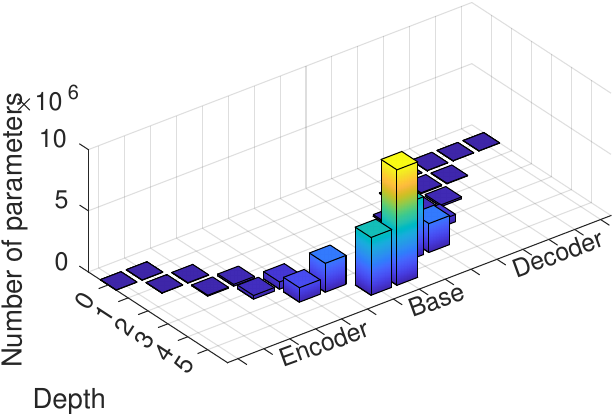}
\caption{Number of parameters.}
\label{fig:3dbarNumParams}
\end{subfigure}
\caption{U-Net computational complexity per $conv2D$ and per transposed $conv2D$ layers.}
\label{fig:staticAnalysis}
\end{figure}

Figure \ref{fig:3dbarNumFLOPS} shows the number of FLOPS for each $conv2D$ layer in the model.
It can be observed that the decoder branch requires roughly twice as many FLOPS as the encoder branch.
Additionally, at each depth level, the second convolution block has an equal number of FLOPS in both the encoder and decoder branches.

The same reasoning applies to the calculation of the number of parameters, since non-convolutional layers represent only 0.08\% of the total.
The number of parameters in the 3x3 $conv2D$ layers mainly depends on the number of filters and the spectral size, both of which increase with the network’s depth.
Consequently, the layers with the highest number of parameters are located near the base of the U-Net.
Equation \ref{equ:paramsTotal} represents the approximate number of parameters of the whole model

\begin{equation}
    params \approx \sum_{j=0}^{n} params_{conv_j} = \sum_{j=0}^{n} o_{f_{j}} \cdot k_{h_{j}} \cdot k_{w_{j}} \cdot i_{c_{j}}
    \label{equ:paramsTotal}
\end{equation}

\noindent where ${}_j$ is the subscript indicating the number of the convolution layer $o_{f}$ is the number of filters or depth of the output feature map, $k_{h}$ is the height of the convolution kernel, $k_{w}$ is the width of the convolution kernel, $i_{c}$ is the number of input channels or depth of the convolution kernel.

Based on Figures \ref{fig:3dbarNumFLOPS} and \ref{fig:3dbarNumParams} and depending on the specific optimization objective used for pruning (whether reducing parameters, FLOPS, a combination of both, or balancing memory accesses against computations \cite{wu2018shift}) it can be determined where the pruning algorithm should focus its efforts.

\subsubsection{Dynamic Analysis}
The basis of the dynamic analysis is a sensitivity analysis, which is a concept that dates back to the optimal brain damage method proposed by \cite{NIPS1989_6c9882bb}.
It is also used in \cite{li2017pruningfiltersefficientconvnets}, one of the first papers that focuses on structured pruning in convolutional networks.

Building on the static analysis presented in Section \ref{sec:staticAnalysis}, the sensitivity analysis is conducted by defining FLOPS reduction as the pruning objective and using the smallest L1 norm as the pruning criterion to select which channels to remove within each layer.
Each layer is then pruned incrementally, applying a \textit{pr} from 0 to 0.9 in steps of 0.1, while the rest of the model is kept frozen.
The final $conv2D$ layer is excluded from pruning since its number of filters must match the number of segmentation classes.
The pruned model is then evaluated according to a given metric such as the IoU.
The tool used for both the sensitivity analysis and the iterative pruning process is the VAI Optimizer 3.5 from AMD-Xilinx, which was made open source in VAI 3.5 \cite{vitisAI}.

\begin{figure}[h!]
\centering
\begin{subfigure}{0.33\linewidth}
\centering
\includegraphics[width=4.6cm]{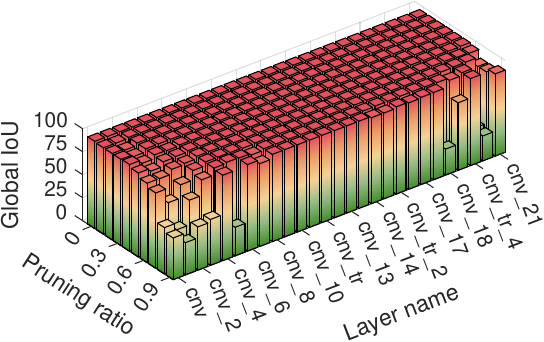}
\caption{Global IoU.}
\label{fig:sensitivity05global}
\end{subfigure}%
\begin{subfigure}{0.33\linewidth}
\centering
\includegraphics[width=4.6cm]{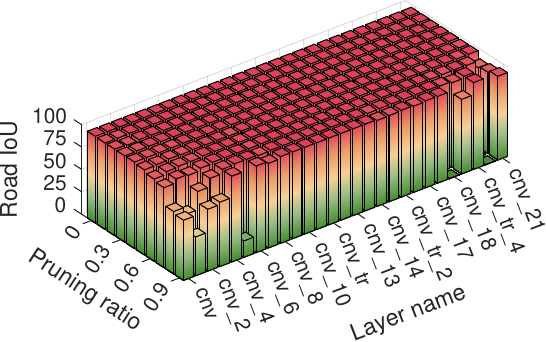}
\caption{Road IoU.}
\label{fig:sensitivity05road}
\end{subfigure}%
\begin{subfigure}{0.33\linewidth}
\centering
\includegraphics[width=4.6cm]{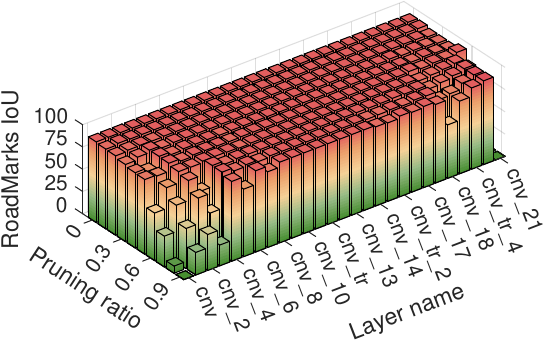}
\caption{Marks IoU.}
\label{fig:sensitivity05roadMarks}
\end{subfigure}\\[1ex]
\begin{subfigure}{0.33\linewidth}
\centering
\includegraphics[width=4.6cm]{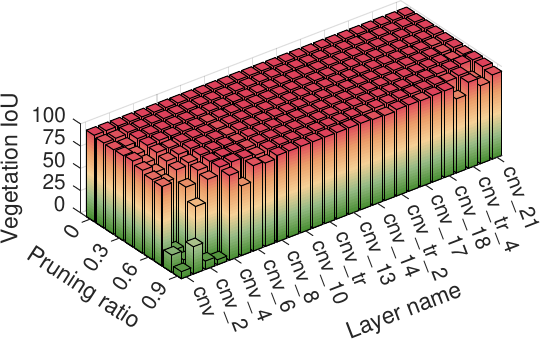}
\caption{Vegetation IoU.}
\label{fig:sensitivity05vegetation}
\end{subfigure}%
\begin{subfigure}{0.33\linewidth}
\centering
\includegraphics[width=4.6cm]{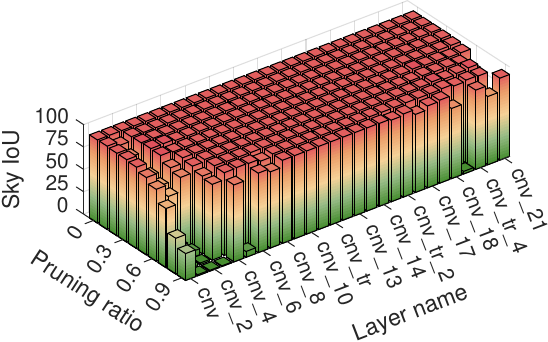}
\caption{Sky IoU.}
\label{fig:sensitivity05sky}
\end{subfigure}%
\begin{subfigure}{0.33\linewidth}
\centering
\includegraphics[width=4.6cm]{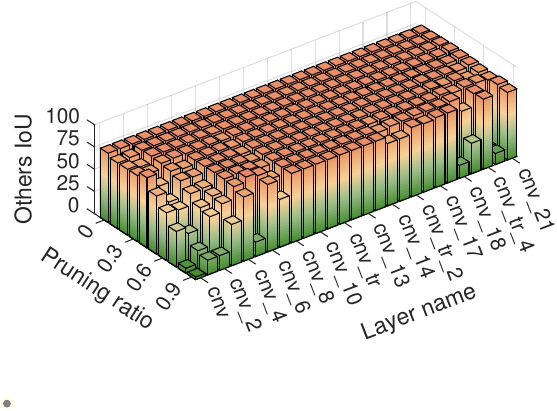}
\caption{Others IoU.}
\label{fig:sensitivity05others}
\end{subfigure}
\caption{Class IoU-based sensitivity analysis on the U-Net $conv2D$ (cnv) and transposed $conv2D$ (cnv\_tr) layers.}
\label{fig:sensitivity05}
\end{figure}

Figures \ref{fig:sensitivity05global} to \ref{fig:sensitivity05others} provide further insight into how the segmentation performance for specific classes is affected by pruning certain filters.
For example, the Road class is generally the most robust; however, for some layers, applying aggressive pruning leads to catastrophic segmentation failure.
The Marks class is particularly sensitive to pruning in the initial layers.
For the Vegetation class, there are significant differences between the most aggressive consecutive \textit{pr}s, but overall, its robustness remains high.
The Sky class shows a particular degradation when the second convolution is pruned, while the behaviour of the Other class resembles that of Marks.
This could be related to the fact that the first two convolution layers focus on the borders and shapes of elements in the images, which are more present in these two classes.
To maintain equal weighting across all classes, wIoU is used as the pruning evaluation metric.

\subsubsection{Iterative Pruning}
The success of the iterative pruning methodology (Algorithm \ref{alg:iterativePruning}), is tightly coupled with a proper overall \textit{pr} initialisation, which indicates the desired level of complexity reduction.
That ratio sets the target FLOPS, a parameter that is fed to the constrained binary search algorithm.
The recursive algorithm determines an optimal, layer-specific pruning scheme (that is, the individual \textit{pr} for each DNN layer) to achieve the desired reduction in FLOPs.
Simultaneously, the process evaluates prunable groups of DNN layers to ensure that pruning does not lead to excessive degradation in performance, with those layers exceeding the user-defined threshold being excluded from the pruning process.
This scheme safeguards the integrity of the model while enabling efficient optimization.

The layer-specific pruning scheme assessment involves taking the following considerations into account: sensitivity of selected layer/ratio pairs, number of locked layers (layers with a value of 0.9 in the pruning scheme) and, optionally, assessment of the IoU after finetuning.
Firstly, if the proposed \textit{pr} for a given layer is related to a low IoU, it is indicative of an IoU degradation for the whole model after finishing the process.
Secondly, if the number of locked layers is high, it is indicative of a possible IoU degradation but, above all, of an IoU degradation in the following iterations.
The fact of being locked suggests that those layers are robust and that they could have been further pruned instead of the non-locked less robust layers.
Finally, optionally, although models from different schemes could be fine-tuned and their IoU scores compared, the main goal of iterative pruning is not only to improve the present results but to prepare the model for further pruning in future iterations.
Once the ideal layer-specific pruning scheme has been selected, finetuning process is performed so as to, at least, partially recover the lost IoU.
From there on, there are two options: either increase the initial \textit{pr} (if the previous approach is deemed conservative) or continue with the following iterations, for which the entire process must be repeated, selecting a new initial \textit{pr}.

\begin{minipage}{0.32\textwidth}
    \begin{algorithm}[H]
    \setcounter{algorithm}{0}
        \footnotesize
        \caption{\footnotesize{Part 1: Do analyses.}}
        \begin{algorithmic}[1]
            \STATE \textbf{for} num\_iters = 1 to N \textbf{do}
            \STATE \quad Static $\rightarrow$ Objective = FLOPS
            \STATE \quad Dynamic $\rightarrow$ Overall \textit{pr} = 0.5
            \STATE \quad \textbf{while} pr $\neq$ $\sum$(pr\_spec) \textbf{do}
            \STATE \quad\quad pr\_spec(FLOPS, 0.5)
            \STATE \quad \textbf{end while}
            \STATE \quad Go to Part 2
            \STATE \textbf{end for}
        \end{algorithmic}
    \end{algorithm}
\end{minipage}%
\hfill
\begin{minipage}{0.32\textwidth}
    \begin{algorithm}[H]
    \setcounter{algorithm}{0}
        \footnotesize
        \caption{\footnotesize{Part 2: Check last iter.}}
        \begin{algorithmic}[1]
            \STATE \textbf{if} num\_iters = N \textbf{then}
            \STATE \quad \textbf{if} $\Delta$wIoU $<$ 1 \textbf{then}
            \STATE \quad\quad Finetune
            \STATE \quad \textbf{else}
            \STATE \quad\quad Go to Part 1 (line 3)
            \STATE \quad \textbf{end if}
            \STATE \textbf{else}
            \STATE \quad Go to Part 3
            \STATE \textbf{end if}

        \end{algorithmic}
    \end{algorithm}
\end{minipage}%
\hfill
\begin{minipage}{0.32\textwidth}
    \begin{algorithm}[H]
    \setcounter{algorithm}{0}
        \footnotesize
        \caption{\footnotesize{Part 3: Other iters.}}
        \begin{algorithmic}[1]
            \STATE \textbf{if} layer $\Delta$wIoU $<$ 0.25 \textbf{and} lock\_layers $<$ 25\% \textbf{and} $\Delta$wIoU $<$ 1 \textbf{then}
            \STATE \quad Finetune
            \STATE \quad Go to Part 1 (line 2)
            \STATE \textbf{else}
            \STATE \quad Go to Part 1 (line 3)
            \STATE \textbf{end if}
        \end{algorithmic}
    \end{algorithm}
\end{minipage}

\begin{center}
    \setcounter{algorithm}{0}
    \captionof{algorithm}{Pseudocode of the Iterative pruning methodology.
    These values are valid for this specific case but may be adjusted depending on application constraints.}
    \label{alg:iterativePruning}
\end{center}

\begin{figure}[h!]
\centering
\begin{subfigure}{0.48\linewidth}
\centering
\includegraphics[width=6cm]{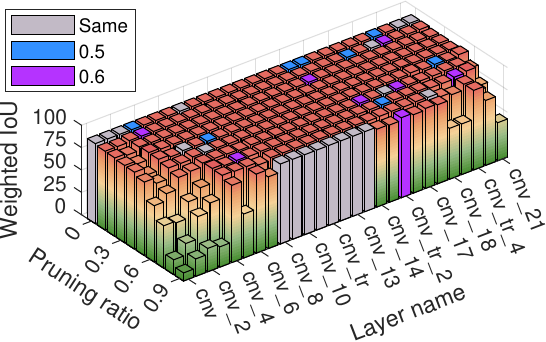}
\caption{Non-pruned model.}
\label{fig:sensitivity0506Sameweighted}
\end{subfigure}
\begin{subfigure}{0.48\linewidth}
\centering
\includegraphics[width=6cm]{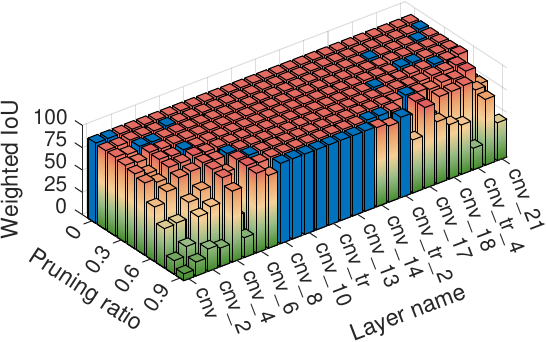}
\caption{One-time 0.5 pruned model.}
\label{fig:sensitivity0505weighted}
\end{subfigure}
\caption{Weighted IoU-based sensitivity analyses of different U-Net $conv2D$ layers.
Homogeneously coloured bars are part of the same layer-specific pruning scheme to achieve a certain overall \textit{pr}.}
\label{fig:iterativePruning}
\end{figure}

The outcome of applying this methodology to this case study can be seen in Figure \ref{fig:sensitivity0506Sameweighted}, which depicts the sensitivity analysis of the non-compressed U-Net utilizing wIoU as the verification metric.
The coloured bars illustrate the layer-specific pruning scheme to achieve overall \textit{pr}s of 0.5 (blue) and 0.6 (purple) in terms of FLOPS.
The grey colour represents the layers that are pruned equally across both pruning schemes.

For an initial \textit{pr} of 0.5, the DNN has been compressed from 34.87G to 16.82G FLOPS and from 31.10 M to 2.48 M parameters (see first row of Table \ref{tab:pruningAnalysis}).
The huge decrease in the number of parameters can be explained by combining Figures \ref{fig:3dbarNumParams} and \ref{fig:sensitivity0506Sameweighted}.
The most pruned layers are the ones that contain the greatest amount of parameters.
The U-Net was then finetuned for a total of 60 epochs with a learning rate of $10^{-6}$ and quantized as described in Subsection \ref{sec:quantization}.
The test results, where the pruned model unexpectedly outperforms the non-pruned model (see first row of Table \ref{tab:postTrainingQuantization}), suggest that further compression could be possible without affecting performance.
To continue with the pruning process, a higher overall \textit{pr} to the base model or a small second \textit{pr} to the pruned model could be applied.

In the first case, a higher overall \textit{pr} value is applied, which causes more layers to lock (see Figure \ref{fig:sensitivity0506Sameweighted}).
Those layers are robust and computationally burdensome, so it could be interesting not to have them locked.
Nevertheless, for the sake of comparison, the base model was also pruned with 0.6, 0.7 and 0.8 \textit{pr}s and finetuned afterwards.
The results of the 0.6-pruned and 0.7-pruned models (see second and third row in Table \ref{tab:pruningAnalysis}) are satisfactory, while the metrics for the 0.8-pruned model (see fourth row in Table \ref{tab:pruningAnalysis}) greatly degraded, especially when looking at the most under-represented classes, which mostly affect wIoU.

\begin{table*}[!h]
\centering
\caption{IoU of the 8-bit quantized U-Net models according to \textit{pr}.}
\label{tab:pruningAnalysis}
\resizebox{13.5cm}{!}{%
\begin{tabular}{ccccccccccc}
\toprule
\textbf{Pruning ratio}  & \textbf{Mparams}  & \textbf{GFLOPS$^{\mathrm{a}}$} & \textbf{Size (MB)} & \textbf{Road} & \textbf{Marks} & \textbf{Vegetation} & \textbf{Sky} & \textbf{Other} & \textbf{Global} & \textbf{Weighted} \\
\midrule
0.5						& 2.48					  & 16.82                    & 2.37               & 97.87         & 88.37               & 93.82               & 92.74        & 78.81          & 94.71           & 88.77             \\
0.6						& 1.62					  & 13.84                    & 1.54               & 97.87         & 87.97               & 93.67               & 92.44        & 78.32          & 94.61           & 88.41             \\
0.7						& 1.01					  & 10.53                    & 0.96               & 97.77         & 87.99               & 93.74               & 92.93        & 77.48          & 94.51           & 88.44             \\
0.8						& 0.59					  & 6.97                     & 0.56               & 96.56         & 80.96               & 93.25               & 88.69        & 75.39          & 93.04           & 83.51             \\
0.75 (0.5 \& 0.5)		& 0.32					  & 8.49                     & 0.31               & 97.72         & 88.15               & 93.85               & 92.45        & 77.38          & 94.47           & 88.37             \\
0.8 (0.5 \& 0.6)		& 0.26					  & 6.62                     & 0.25               & 93.39         & 85.92               & 92.48               & 90.83        & 62.42          & 89.94           & 84.43             \\
0$^{\mathrm{b}}$		& 7.76					  & 31.76$^{\mathrm{c}}$     & 7.41               & 96.93         & 80.28               & 94.04               & 91.54        & 73.48          & 93.38           & 83.45             \\
\bottomrule
\multicolumn{10}{l}{$^{\mathrm{a}}$ Input size of 192x384.
$^{\mathrm{b}}$ This row corresponds to the floating-point model of depth 4 used in \cite{icecs2023}.} \\
\multicolumn{10}{l}{$^{\mathrm{c}}$As a consequence of the model depth, the input was adjusted to 208x400 pixels.}
\end{tabular}}
\end{table*}

To address this issue, a class-aware pruning strategy could be employed.
This could involve focusing on the sensitivity analyses of the specific classes (see Figure \ref{fig:sensitivity05}) and adjusting the \textit{pr} of critical layers accordingly.
Alternatively, a modified wIoU metric could be introduced, assigning higher importance to sensitive classes to preserve their predictive performance.
However, despite the inherent parallelism and overparameterization of DNNs, they are not entirely immune to its effects.
An aggressive pruning ratio of 0.8 may still hinder the DNN’s ability to preserve essential feature representations of those classes, regardless of the post-training pruning methodology applied.

\begin{figure}[h!]
\centering
\includegraphics[width=7.5cm]{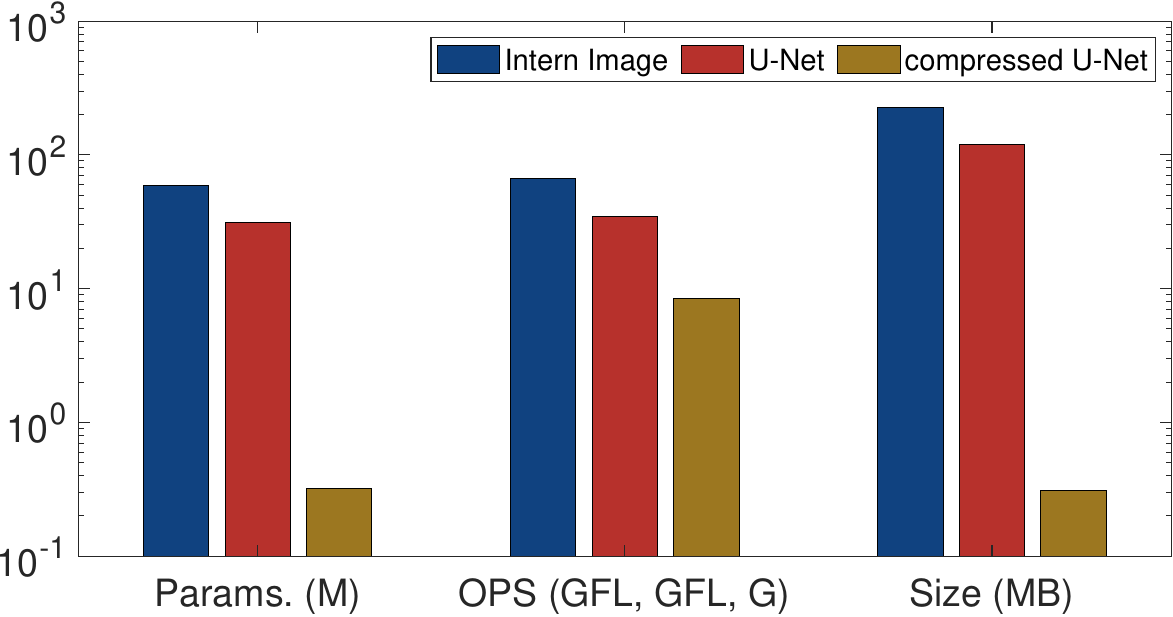}
\caption{Number of parameters (left), operations (centre) and size (right) comparison among Intern Image (blue bar), U-Net (red bar) and the compressed U-Net (brown bar).}
\label{fig:SotAcomparison}
\end{figure}

In the second case, a second iteration is performed, requiring the process to be repeated and a new wIoU-based sensitivity analysis to be conducted (see Figure \ref{fig:sensitivity0505weighted}).
The initial \textit{pr} is selected based on previous results.
A second \textit{pr} of 0.4 resulted in an overall \textit{pr} of 0.7, yielding good results.
A second \textit{pr} of 0.6 gave an overall \textit{pr} of 0.8, which did not produce favourable outcomes.
Therefore, a second \textit{pr} of 0.5 is chosen, leading to an overall \textit{pr} of 0.75.
After finetuning the model, the loss in wIoU compared to the original and the 0.5-pruned models is below 0.5 IoU points (see fifth row in Table \ref{tab:pruningAnalysis}) while a reduction of 75\% in OPS and of 99\% in parameters is achieved.
All in all, the applied model optimization methodology (combining post-training iterative pruning and quantization) has produced an architecture that maintains system accuracy while reducing model complexity by orders of magnitude, as shown in Figure \ref{fig:SotAcomparison}.

It is also important to note that, as expected, training shallower dense U-Net models (depths 2, 3, and 4) with fewer initial convolutional filters (8, 16, 32) resulted in considerably worse performance (about 5 IoU points lower) compared to pruning trained large sparse models (see seventh row in Table \ref{tab:pruningAnalysis}).
In fact, that poor result is very similar to one of the worst model of the one-time pruning process (ratio of 0.8), but with 13x parameters and 4x GFLOPS.

\subsubsection{Pruning Efficiency: Iterative vs. Non-Iterative Approaches}
This section compares the pruning efficiency at a global pruning ratio of 0.8, evaluating both iterative pruning (0.5 \& 0.6 steps) and one-time pruning (0.8) methods, along with their resulting layer-specific pruning schemes.

Regarding the comparison of results, although the gIoU is approximately 3 points lower for the iterative pruning approach, the wIoU (the primary focus) is nearly 1 point higher (see the fourth and sixth rows of Table \ref{tab:pruningAnalysis}).
Moreover, the iterative pruning process allows previously locked layers to be further pruned, resulting in a model with roughly half the number of parameters compared to the non-iterative approach.
Despite this difference in parameters, the overall FLOPS remain very similar due to the same overall pruning ratio.

\begin{figure}[h!]
\centering
\includegraphics[width=13.5cm]{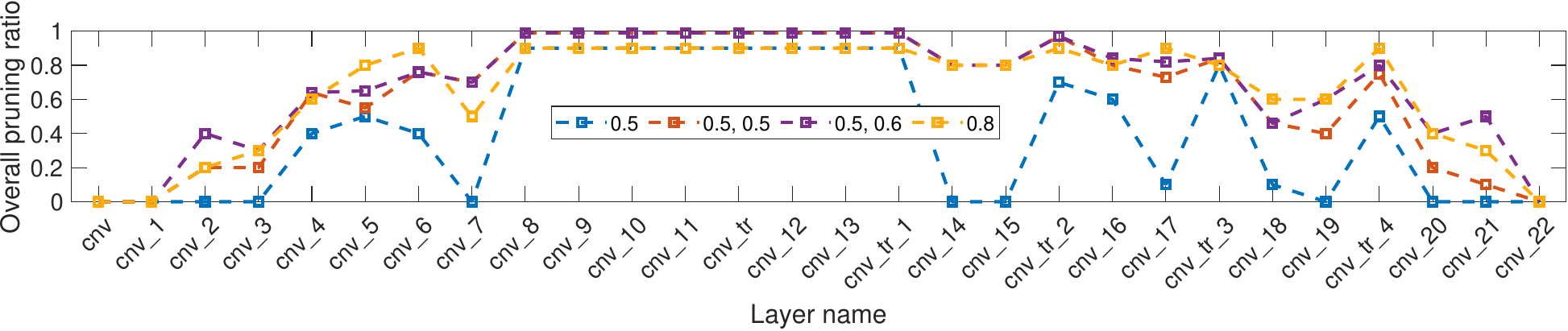}
\caption{Comparison of the final \textit{pr} of: 0.5 (blue), 0.5 \& 0.5 (red), 0.5 \& 0.6 (yellow) and 0.8 (purple) approaches.}
\label{fig:ratioResultante3}
\end{figure}

As for the layer-specific pruning scheme, the purple and yellow lines in Figure \ref{fig:ratioResultante3} represent the resultant \textit{pr} for each layer for the one-time and iterative pruning approaches.
It can be seen how the iterative pruning method allows for a more flexible pruning outcome as the number of locked layers is kept at 8 matching the count from the first iteration (indicated by the blue line in Figure \ref{fig:ratioResultante3}) while the number increases to 12 for the one-time pruning.
In fact, those 8 layers have now been pruned again with a ratio of 0.9 (overall \textit{pr} of 0.99).

\newpage

This trend also holds when applying a second \textit{pr} of 0.5 (red line in Figure \ref{fig:ratioResultante3}).
Comparing the red and yellow lines in Figure \ref{fig:ratioResultante3} reveals which pruned layers caused the severe degradation in results, despite the overall \textit{pr}s are quite similar (0.75 and 0.8 respectively).
The \textit{pr} of the central area is almost the same (from layers $cnv\_6$ to $cnv\_tr\_2$) while the first group of layers and the last ones, the ones that are usually less robust, are pruned differently.
Expectedly, the yellow lines are never below the red lines.

\subsubsection{Pruning Timing: Pre-training vs. Post-training Approach}
As detailed in Section \ref{sec:current_pruning_methods}, previous studies \cite{ditschuneit2022auto, iurada2024finding} have explored pre-training pruning for semantic segmentation.
Although the reported results are not particularly encouraging, pre-training pruning was also investigated to cover a broader range of approaches.
However, since both reported methods rely on ImageNet-pretrained networks and the aim was to explore pruning ratios beyond trivial levels, these methods were not adopted directly.
Instead, the previously mentioned initialization strategy was retained, and the same pruning criteria were applied, exploring overall pruning ratios of [0.6, 0.7, and 0.8].
This resulted in a reduction of at least 95\% in the number of parameters and 60\% in FLOPS, which is considered significant.
However, this model simplification did not lead to the expected reduction in training time (from 8 hours to 6.5 hours), although other factors, such as hardware memory constraints or data size, could have contributed to this outcome.

\begin{figure}[h!]
\centering
\includegraphics[width=13.5cm]{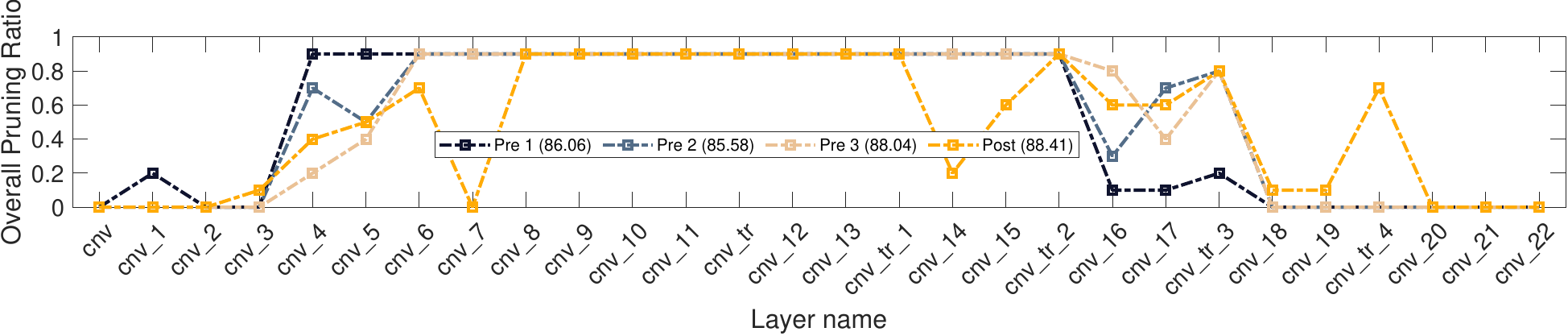}
\caption{Comparison of the pruning schemes of the pre-training approach (black, blue and light brown) and the post-training approach (orange) for a 0.6 overall pruning ratio.}
\label{fig:ACMratio6}
\end{figure}

Figure \ref{fig:ACMratio6} shows the pruning schemes at an overall pruning ratio of 0.6, revealing significant WIoU variation across initializations with the pre-training pruning method.
Initializations with lower WIoU (black and blue lines) tend to over-prune layers $cnv\_4$ and $cnv\_5$ while under-pruning $cnv\_16$ and $cnv\_17$.
In contrast, the best-performing initialization (light brown line) prunes $cnv\_4$ and $cnv\_5$ less aggressively while applying heavier pruning to $cnv\_16$ and $cnv\_17$, aligning with trends observed in post-training pruning (Figure \ref{fig:ratioResultante3} and the orange line in Figure \ref{fig:ACMratio6}) and with modern architectural architectures with complex encoders and lightweight decoders \cite{papadeas2021real}.

This behaviour is also influenced by the model’s structure, where the decoder requires more FLOPS than the encoder (Figure \ref{fig:3dbarNumFLOPS}).
In the post-training scheme, several pruning peaks align with the transposed convolution layers responsible for upsampling decoder activations, indicating increased reliance on encoder information.
Additionally, the minimal pruning of layers $cnv\_7$ and $cnv\_14$ reflects their critical role in skip connections, which directly transmit features from the encoder to the decoder.

A key distinction between the pre-training and post-training pruning schemes is the flatter pruning distribution observed in the pre-training case.
In particular, layers $cnv\_6$, $cnv\_7$, $cnv\_14$, and $cnv\_15$ are pruned much more aggressively under pre-training pruning.
Such pronounced variations in pruning across layers are more likely to hinder feature extraction and, consequently, the segmentation accuracy.

\begin{figure}[h!]
\centering
\includegraphics[width=13.5cm]{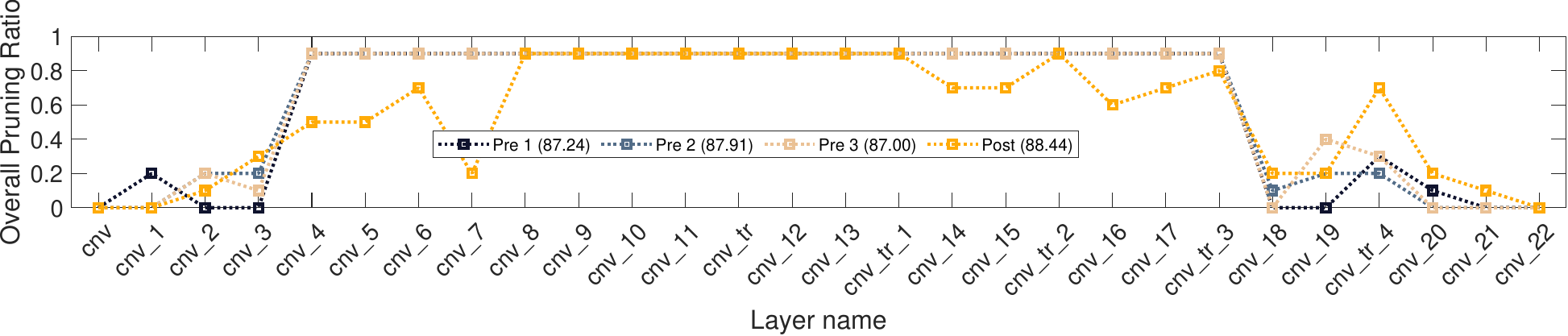}
\caption{Comparison of the pruning schemes of the pre-training approach (black, blue and light brown) and the post-training approach (orange) for a 0.7 overall pruning ratio.}
\label{fig:ACMratio7}
\end{figure}

Figure \ref{fig:ACMratio7} shows the pruning schemes at a pruning ratio of 0.7, which exhibit the highest similarity and lowest variance across initializations.
In the pre-training case, the pruning distribution remains relatively flat around the base of the model, with heavier pruning concentrated in the decoder, particularly at layers $cnv\_tr\_3$ and $cnv\_tr\_4$.
Likewise, in the post-training scheme, the decoder layers show the most significant pruning increase compared to the 0.6 ratio case.

\begin{figure}[h!]
\centering
\includegraphics[width=13.5cm]{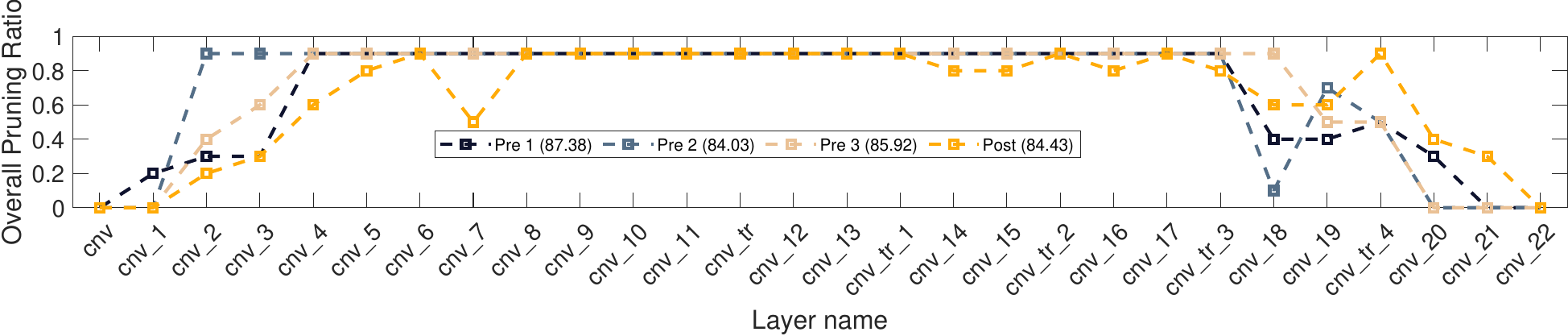}
\caption{Comparison of the pruning schemes of the pre-training approach (black, blue and light brown) and the post-training approach (orange) for a 0.8 overall pruning ratio.}
\label{fig:ACMratio8}
\end{figure}

Finally, Figure \ref{fig:ACMratio8} illustrates the pruning schemes at a 0.8 pruning ratio, again showing significant differences across initializations and the highest variability in WIoU outcomes.
As observed previously, initializations (blue line) that apply more aggressive pruning to both the encoder ($cnv\_2$ and $cnv\_3$) and the decoder ($cnv\_18$ and $cnv\_19$) tend to result in lower WIoU scores compared to more balanced pruning strategies, such as the ones represented by the black and light brown lines.

Despite the increased overall pruning ratio, the post-training pruning strategy tends to preserve certain internal layers ($cnv\_7$, $cnv\_14$, $cnv\_15$, and $cnv\_16$) while applying more aggressive pruning to the outer layers ($cnv\_18$, $cnv\_19$, $cnv\_20$, and $cnv\_21$), leading to performance degradation.
Notably, this is the only configuration where pre-training pruning outperforms post-training pruning (in two initializations).
However, this result is not entirely unexpected, given that, despite the high pruning ratio, the pre-trained pruned models still retain 0.41 M and 0.38 M parameters, respectively, compared to 0.26 M parameters in the post-training pruned model (Table \ref{tab:pruningAnalysis}).

In conclusion, both pre-training and post-training pruning approaches have been comprehensively evaluated in terms of training time and WIoU.
The iterative post-training pruning method consistently outperformed pre-training pruning.
In the pre-training approach, the highest WIoU scores were 88.04 at a pruning ratio of 0.6, 87.91 at 0.7, and 87.38 at 0.8, all below the results of the iterative post-training method (88.37 at 0.75) and the original unpruned network (88.40)
Notably, for pruning ratios of 0.6 and 0.7, pre-training pruning achieved lower scores than post-training pruning, despite the latter not being applied iteratively.
Only at a 0.8 pruning ratio did pre-training pruning outperform the iterative method in two of three initializations, likely due to the relatively high number of parameters retained in those models.

These results suggest that it could be worth exploring asymmetric encoder-decoder architectures with deeper encoder branches and lighter decoder even though, as discussed in Section \ref{sec:sota}, the recent hyperspectral benchmark HyperSeg \cite{theisen2024hs3} achieved better results on the same dataset using a U-Net compared to DeepLabv3+ \cite{chen2018encoder}, despite the latter having a more complex encoder and significantly simplified decoder.

\section{Integrating Raw Data Preprocessing in DNN Co-design}\label{sec:rawImagePreprocessing}
\begin{figure}[h!]
\centering
\includegraphics[width=13cm]{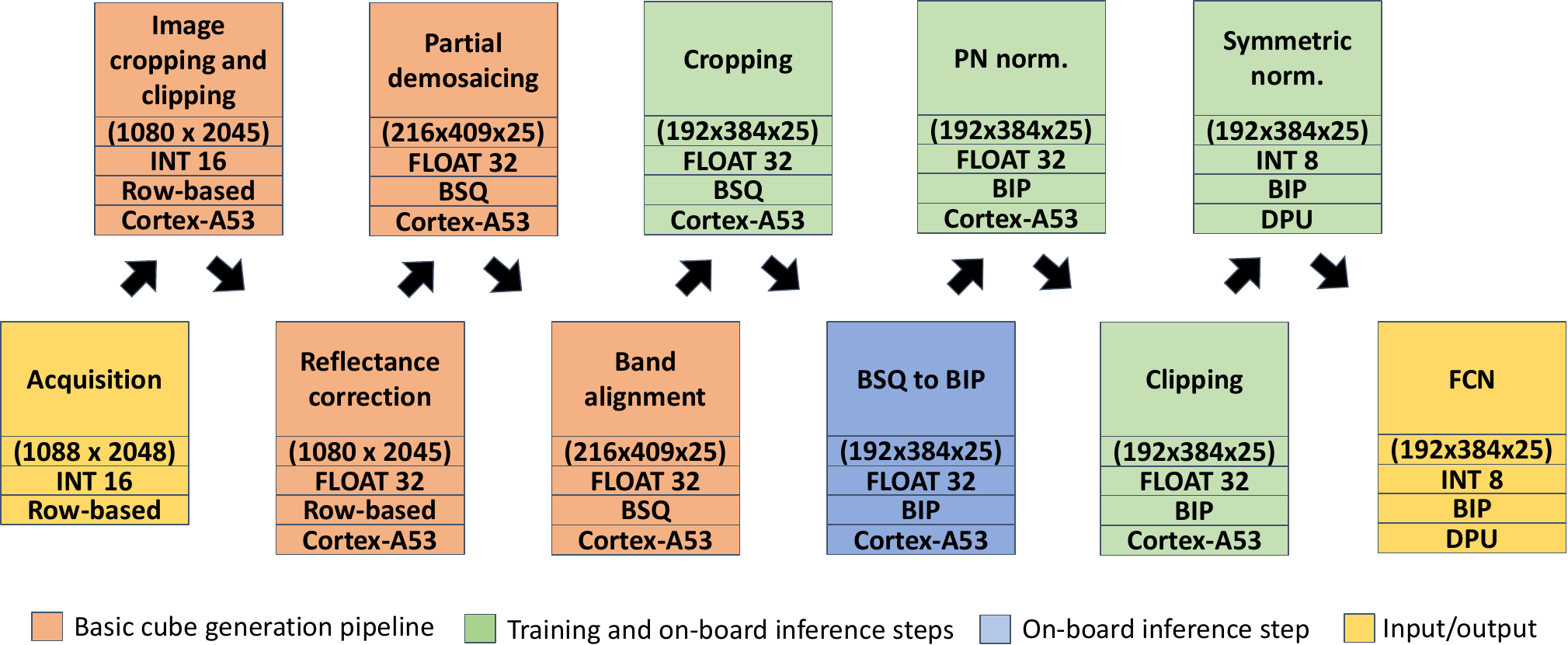}
\caption{Pipeline scheme comprising from raw image preprocessing to DNN inference.
In each step, information about image size, data representation, memory arrangement and task distribution is provided.}
\label{fig:cube_generation}
\end{figure}

The basic preprocessing of HSI-Drive v2.0 raw images (see orange boxes in Figure \ref{fig:cube_generation}) includes the following stages: cropping and clipping, reflectance correction through dark and flat images, partial demosaicing and band alignment via spatial bilinear interpolation (see \cite{gutierrez2023chip} for a detailed description).

Additionally, to adapt the cube for the U-Net architecture, an extra cropping or padding step is required due to the mismatch between the cube's spatial size (216x409) and the DNN encoder-decoder architecture with a depth of 5 (see Section \ref{sec:modelArchitectureTraining}).
The spatial size must be a multiple of $2^{depth\_level}$, so the nearest valid sizes are 192x384 (requiring cropping) and 224x416 (requiring padding).
Padding could introduce issues with pixel values at the edges, which would hinder efficient preprocessing.
Given these considerations and the suitability of 192x384 resolution for the application, cropping has been chosen.
Notably, this cropping step cannot be combined with the initial one, as the edge pixels resulting from the band alignment process require data that would be missing if cropped earlier.

Apart from that, applying per-pixel normalization (PN) is recommended to generate a cube robust to varying lighting conditions.
However, applying PN alters the data distribution, concentrating most of the values around 0.04 (the inverse of 25, the spectral size) and making the interval [0, 0.08] contain 99.72\% of all pixel values (see \cite{icecs2023}).
If the DNN is to be quantized, applying image clipping is then beneficial.
Clipping according to the channel distribution accurately represents 99.95\% of the data, saving 3 integer bits to enhance quantization process resolution.
Finally, symmetric normalization is applied during training, accelerated by hardware (see Section \ref{sec:hardwareAcceleratedSteps}), and used as the DNN input.
The full pipeline, including data sizes, data representations, and ARM/DPU task distribution, is shown in Figure \ref{fig:cube_generation}.

\begin{figure}[h!]
\centering
\begin{subfigure}{0.53\linewidth}
\centering
\includegraphics[width=8cm]{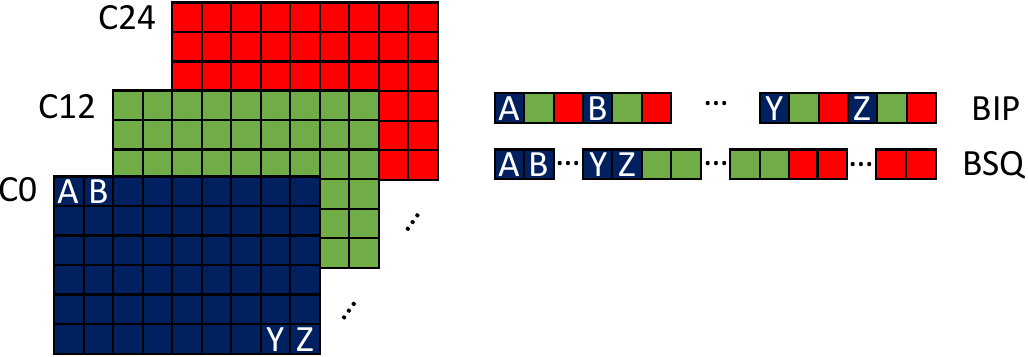}
\caption{Band interleaved by pixel (BIP) and band sequential format (BSQ) data organizations.}
\label{fig:bip_bil_bsq}
\end{subfigure}
\begin{subfigure}{0.46\linewidth}
\centering
\includegraphics[width=6cm]{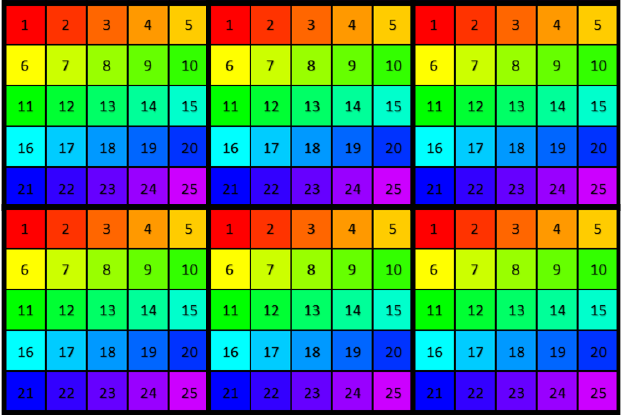}
\caption{A 10x15 frame containing six 5x5 tiles.}
\label{fig:msfa}
\end{subfigure}
\caption{Comparison of cube data organization in memory (left).
The hyperspectral Fabry-Perot filter array used during raw-image acquisition (right).}
\label{fig:cubeAdaptation}
\end{figure}

Several constraints impact the efficiency of data representation.
First, the camera outputs a 1088x2048 frame with 5x5 tiles in row-major order (see Figure \ref{fig:msfa}).
Second, the AMD-Xilinx DPUCZDX8G inference engine requires data in Band Interleaved by Pixel (BIP) format.
Third, cropping must be applied after band alignment to preserve edge spatial information.
Finally, some operations are channel-wise (e.g., band alignment, cropping), while others are pixel-wise (e.g., PN norm, clipping, symmetric norm).

Considering these constraints, converting raw data to Band Sequential (BSQ) format is more efficient initially since BSQ stores each band sequentially, matching the row-major order and improving memory access.
Channel-wise operations benefit from BSQ’s consecutive spatial storage, whereas pixel-wise operations are faster in BIP, which stores spectral data contiguously (Figure \ref{fig:bip_bil_bsq}).
As shown in Figure \ref{fig:cube_generation}, BSQ-favourable steps are performed early in the pipeline to optimize data transfer and speed, while BIP-favourable steps occur just before inference.

This approach allows loading entire rows into cache, reducing cache misses and accelerating processing.
In contrast, BIP leads to scattered memory access and lower performance.
Thus, converting the 2D raw image into a 3D BSQ cube enables faster, more efficient processing.
Overall, the most efficient solution is to convert from BSQ to BIP mid-pipeline, as illustrated in Figure \ref{fig:cube_generation}.
The next section details the reordered operations to save computational time.

\section{DNN Deployment and Computational Profiling}\label{sec:DeploymentAndTesting}
\subsection{Hardware Selection}
ADS applications demand low power, cost efficiency, high throughput, and minimal latency.
The AMD-Xilinx KV260 is well-suited for these needs, offering adaptable, high-performance AI processing at the edge, ensuring rapid development and production-ready solutions \cite{k26somIdeal}.
At the core of development board is the K26 SOM, which features a Zynq UltraScale+ MPSoC (see Figure \ref{fig:BlockDiagram}).
This includes a 64-bit quad-core ARM Cortex-A53 processor and a dual-core Cortex-R5F real-time processor within the processing systems, all integrated with a 16nm FinFET Programmable Logic (PL).
The system accesses a 4 GB, 64-bit wide DDR4 external memory with a speed of 2400 Mb/s \cite{k26datasheet}.
The PL can accommodate up to four DPU cores, which utilise high-speed data pathways and parallel processing elements optimized for fixed-point arithmetic units \cite{dpu}.
The heterogeneous architecture of the K26 SOM allows for the optimization of the deployment of AI models by offloading the execution of the corresponding tasks to both parts of the SOM.

The DPU incorporates several on-chip memory banks, including block RAM (BRAM) and UltraRAM, to store weights, biases, and intermediate feature maps.
Each block RAM holds 36Kb and can be configured as 4096x9b, 2048x18b, or 1024x36b, while each UltraRAM provides 288Kb with a fixed configuration of 4096x72b.
UltraRAM can be used to replace BRAM in systems with limited BRAM resources, making it particularly useful in specific hardware implementations, such as the softmax function, which uses four BRAMs \cite{dpu}.

In total, the DPU allocates up to 17 memory banks (256 KB per bank) for storing weights.
Given that the pruned U-Net with 320K 8-bit parameters (including biases) would occupy only two memory banks, there is still potential room for parallel data access by utilizing additional banks.
Additionally, the DPU dedicates one memory bank (32 KB) for biases and up to 24 memory banks (128 KB per bank, for a total of 3 MB) for activations.

DPU architectures are categorized based on their peak operations per cycle, determined by pixel parallelism, input channel parallelism, and output channel parallelism.
Smaller DPUs, which contain fewer DSPs, offer lower computational throughput and fewer weight banks, although they maintain the same number of bias and activation banks as larger configurations (see Table \ref{tab:resourcesDPUversion}).

\begin{table}[h!]
\centering
\caption{Resource usage of the various DPU configurations}
\label{tab:resourcesDPUversion}
\resizebox{8cm}{!}{%
\begin{tabular}{ccccc}
\toprule
\textbf{Parameter / DPU name}         & \textbf{B4096} & \textbf{B3136} & \textbf{B3136} & \textbf{B2304}    \\
\midrule
\textbf{Pixel parallelism}           & 8        & 8         & 8         & 8       \\
\textbf{Input channel parallelism}   & 16       & 14        & 14        & 12      \\
\textbf{Output channel parallelism}  & 16       & 14        & 14        & 12      \\
\textbf{DSP}                         & 710      & 566       & 566       & 438     \\
\textbf{RAM usage}                   & low      & high      & low       & high    \\
\textbf{Number of bias banks}        & 1        & 1         & 1         & 1       \\
\textbf{Number of weight banks}      & 17       & 15        & 15        & 13      \\
\textbf{Number of activation banks}  & 16       & 24        & 16        & 24      \\
\textbf{BRAM}                        & 75       & 130       & 78        & 110     \\
\textbf{URAM}                        & 48       & 40        & 40        & 36      \\
\textbf{LUTs}                        & 50332    & 50007     & 48632     & 45100   \\
\textbf{Flip flops}                  & 99035    & 82505     & 81008     & 71298   \\
\bottomrule
\end{tabular}}
\end{table}

The whole segmentation pipeline was coded in C++ (2017 standard) and cross-compiled for execution as an embedded system within a custom PetaLinux distribution, running on the KV260.
Following the methodology employed in prior studies \cite{gutierrez2023chip, icecs2023}, the performance was optimized by combining data-level parallelism using SIMD techniques \cite{ARM2020Neon} with thread-level parallelism via OpenMP pragmas \cite{dagum1998openmp}, effectively reducing computational latency.

\subsection{Hardware-accelerated Cube Preprocessing Steps}\label{sec:hardwareAcceleratedSteps}
As explained in Section \ref{sec:rawImagePreprocessing}, the last preprocessing step of the raw data is the symmetric normalization, which fits the input cubes to [-1, 1) range.
It can be seen in Equation \ref{equ:symmetricNormalization} that channel-wise symmetric normalization is equivalent to a depthwise convolution operation (weight values are $\frac{2}{max_i-min_i}$ and bias values are $(\frac{2*min_i}{min_i - max_i} - 1)$), so this layer can be inserted between the input layer and the first convolutional layer, enabling its acceleration by the DPU

\begin{equation}
    \hat{x_i} = 2 * \frac{x_i - min_i}{max_i - min_i} - 1 = \frac{2}{max_i-min_i}*x_i + (\frac{2*min_i}{min_i - max_i} - 1)
    \label{equ:symmetricNormalization}
\end{equation}

\noindent where $x_i$ is the unnormalized 192x384 slice of channel $i$, $max_i$ is the maximum value of channel $i$, $min_i$ is the minimum value of channel $i$ and $\hat{x_i}$ is the symmetrically normalized 192x384 slice of spectral $i$.

The main drawback of this methodology is the need to requantize the model.
The quantization input symmetry was changed from symmetric to anti-symmetric because the DNN input now comes from the clipping process output (see Figure \ref{fig:cube_generation}), where data values range from 0 to 0.149 (the highest clipped value).
Since Vitis AI only supports signed quantization, using scaled unsigned inputs reduces the dynamic range by half \cite{vitisAI}.
Therefore, it is essential to evaluate whether this requantization significantly degrades the model’s performance.

\begin{figure}[h!]
\centering
\begin{subfigure}{0.33\linewidth}
\centering
\includegraphics[width=4.5cm]{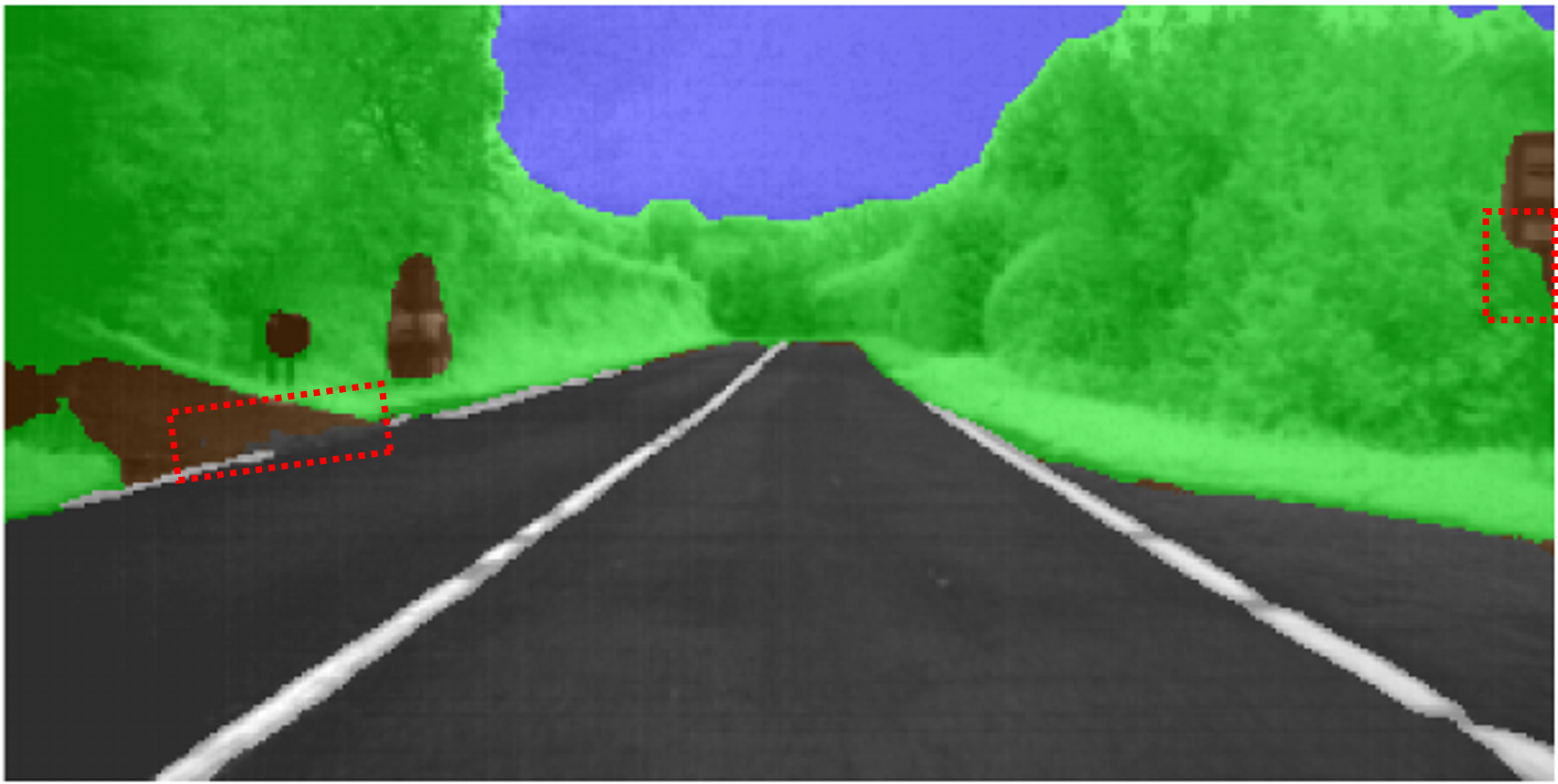}
\caption{$nf1322\_058$.}
\label{fig:dpuOutputSoftware_nf1322_058}
\end{subfigure}%
\begin{subfigure}{0.33\linewidth}
\centering
\includegraphics[width=4.5cm]{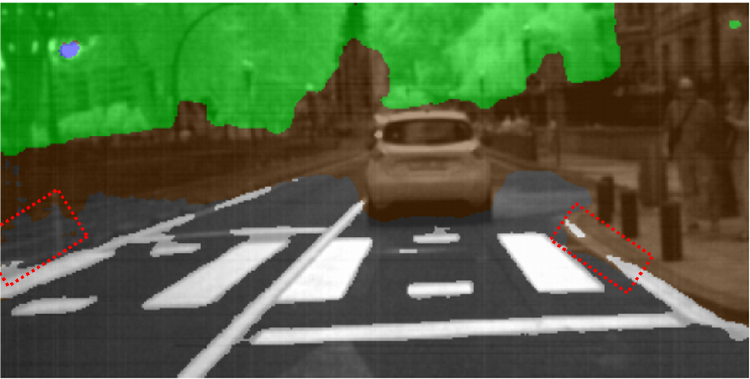}
\caption{$nf2221\_015$.}
\label{fig:dpuOutputSoftware_nf2221_015}
\end{subfigure}%
\begin{subfigure}{0.33\linewidth}
\centering
\includegraphics[width=4.5cm]{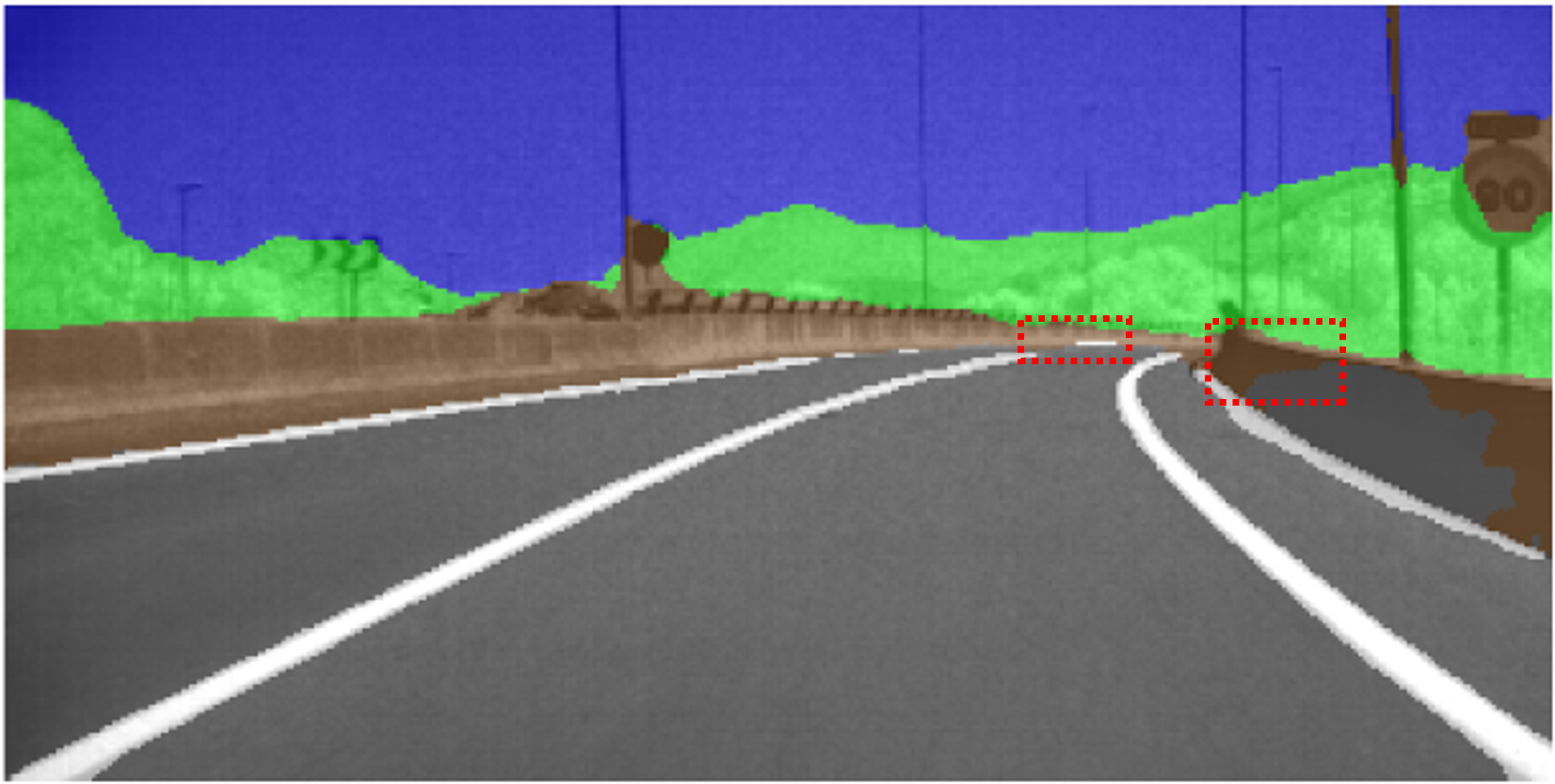}
\caption{$nf3123\_149$.}
\label{fig:dpuOutputSoftware_nf3123_149}
\end{subfigure}
\caption{DPU output when the normalization is performed by software/explicitly.}
\label{fig:dpuOutputSoftware}
\end{figure}

\begin{figure}[h!]
\centering
\begin{subfigure}{0.33\linewidth}
\centering
\includegraphics[width=4.5cm]{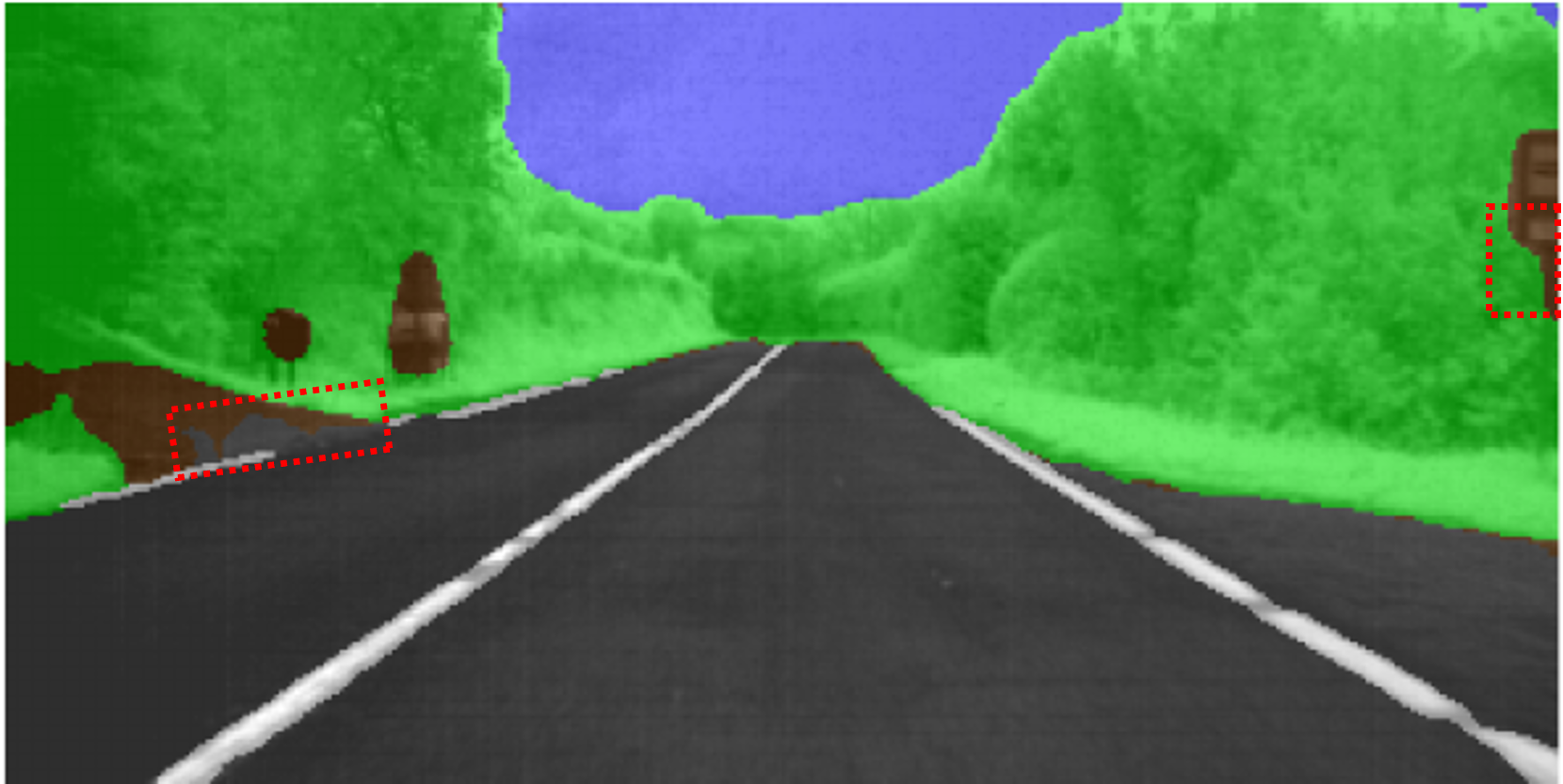}
\caption{$nf1322\_058$.}
\label{fig:dpuOutputHardware_nf1322_058}
\end{subfigure}%
\begin{subfigure}{0.33\linewidth}
\centering
\includegraphics[width=4.5cm]{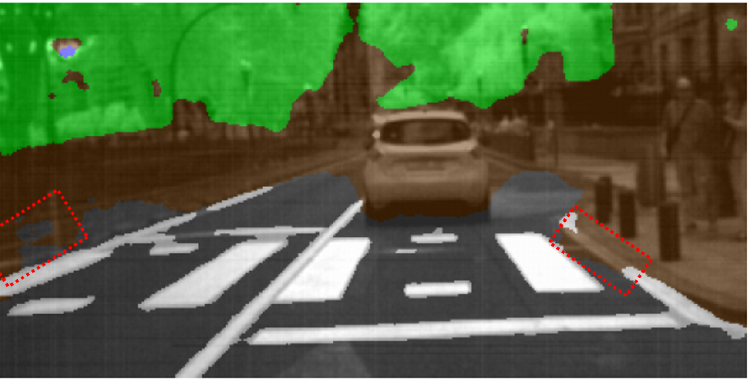}
\caption{$nf2221\_015$.}
\label{fig:dpuOutputHardware_nf2221_015}
\end{subfigure}%
\begin{subfigure}{0.33\linewidth}
\centering
\includegraphics[width=4.5cm]{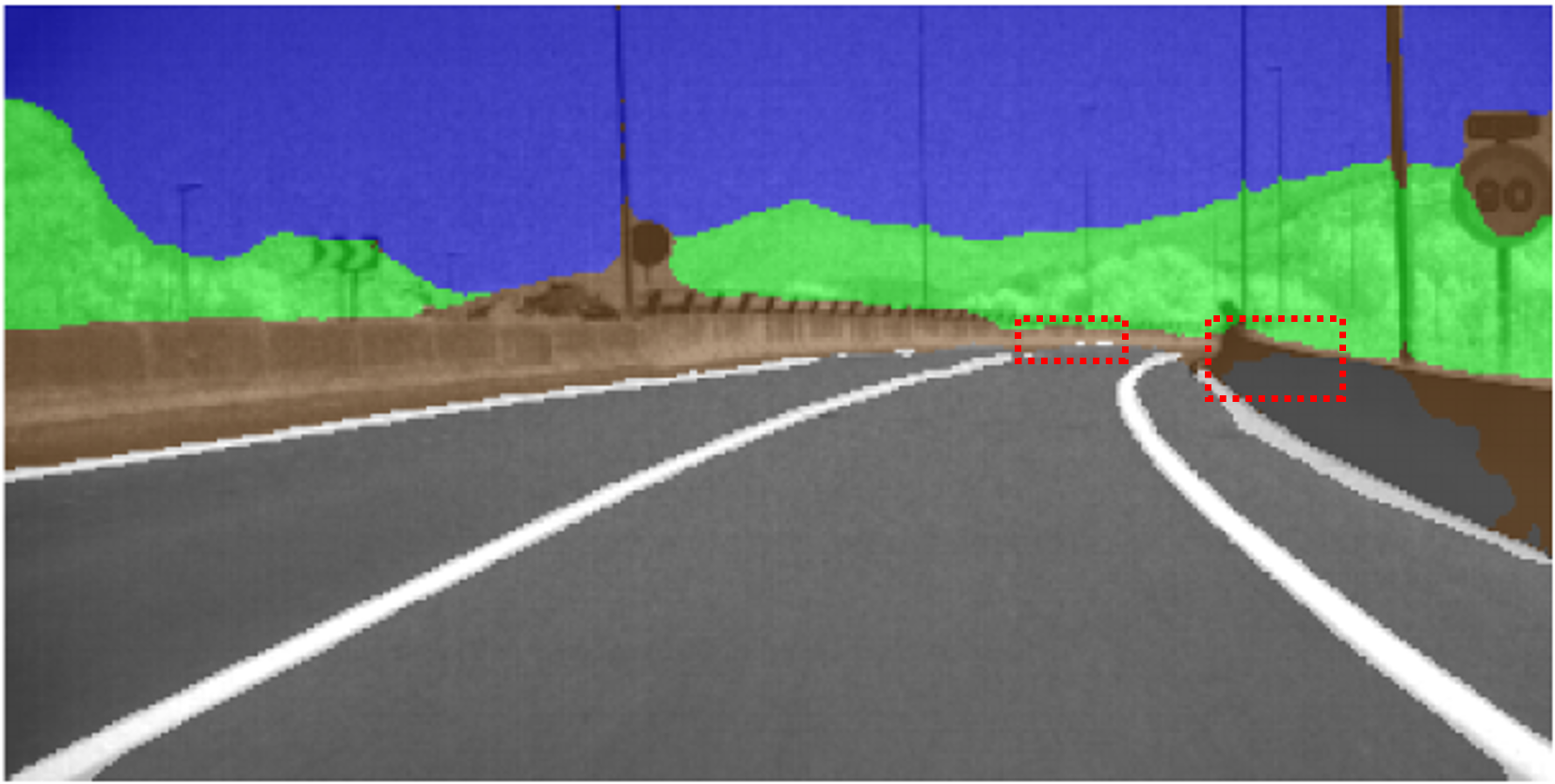}
\caption{$nf3123\_149$.}
\label{fig:dpuOutputHardwarenf3123_149}
\end{subfigure}
\caption{DPU output when the normalization is performed by hardware/implicitly.}
\label{fig:dpuOutputHardware}
\end{figure}

It has been verified that the winning class changes for only about 2\% of all pixels on average.
Moreover, as shown in the comparison between Figures \ref{fig:dpuOutputSoftware} and \ref{fig:dpuOutputHardware}, these pixels are generally located at the boundaries between two or more classes or in regions where surrounding pixels are also misclassified.
Therefore, the impact of the requantization process on overall performance degradation is negligible.

\subsection{Pruning and Its Influence on Deployment Efficiency}
\begin{figure}[h!]
\centering
\begin{subfigure}{0.48\linewidth}
\centering
\includegraphics[width=6.5cm]{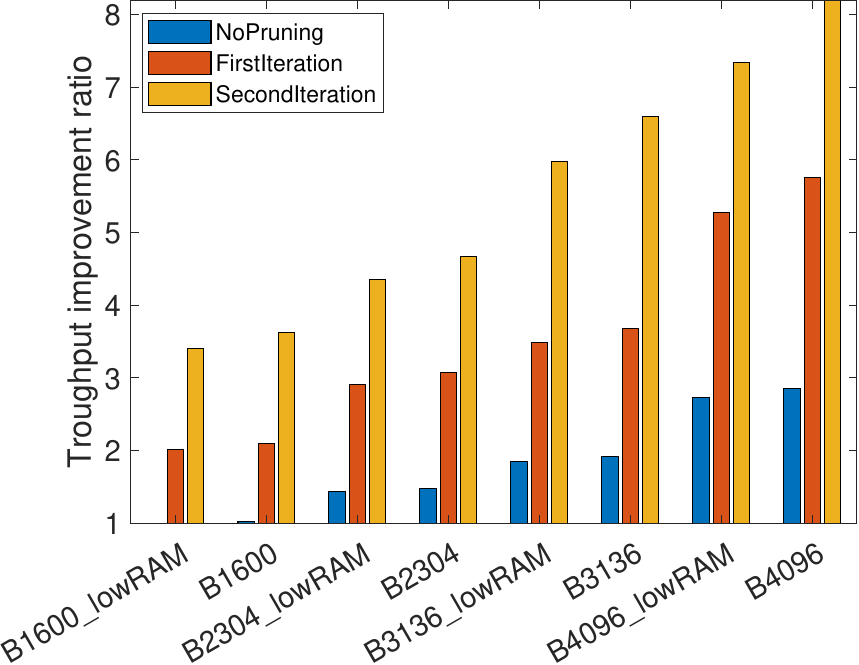}
\caption{Throughput ratio.}
\label{fig:througputRatioBar}
\end{subfigure}
\begin{subfigure}{0.48\linewidth}
\centering
\includegraphics[width=6.5cm]{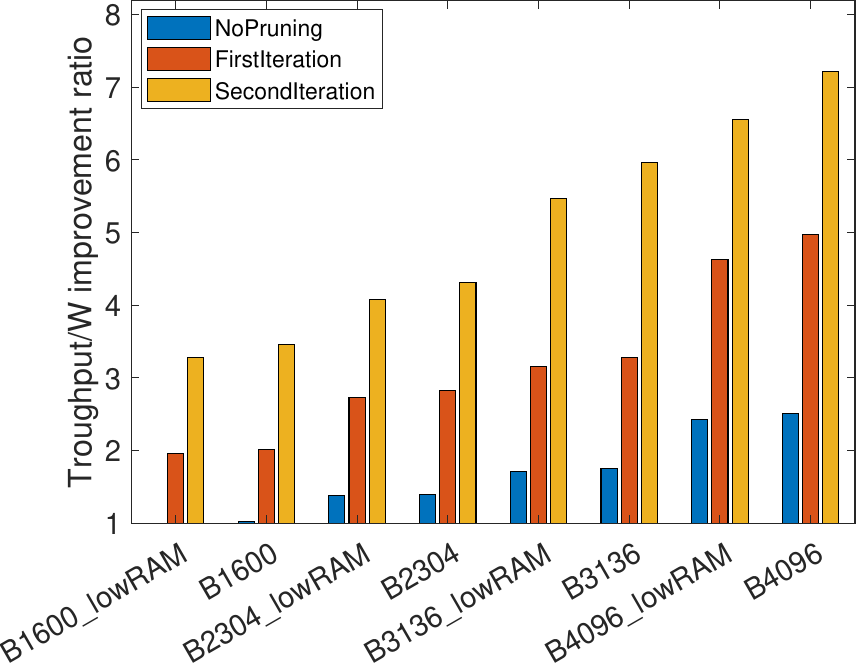}
\caption{Throughput/W ratio.}
\label{fig:througput_WRatioBar}
\end{subfigure}
\caption{Assessment of the impact of pruning iterations and DPU choice on inference throughput (left) and throughput/W (right).}
\label{fig:pruningEffectDPUBenchmark}
\end{figure}

Although pruning significantly reduces model size and complexity, accurately predicting the resulting inference speed-up is challenging due to factors such as preprocessing, quantization, external memory access, and the optimization of the processor’s logical and computational resources.
Therefore, the quantized unpruned, one-time pruned, and two-times pruned models were evaluated in terms of frames per second (Figure \ref{fig:througputRatioBar}) and frames per second per watt (Figure \ref{fig:througput_WRatioBar}) when deployed across various DPU architecture versions.
Smaller versions than the DPU B1600 were excluded, as their inference latency (highest at 74.023 ms for the low RAM, two-times pruned B1600) would exceed the latency of the preprocessing stage.

As shown in Figure \ref{fig:througputRatioBar}, increasing the number of operations per cycle in the DPU correlates with improved throughput.
Similarly, increasing the number of pruning iterations also leads to throughput gains.
Regarding power consumption during inference (Figure \ref{fig:througput_WRatioBar}), the improvement ratio slightly decreases due to a small increase in power usage associated with higher operations per cycle.
However, the difference in power consumption between the pruned models for a given DPU configuration is negligible
The pruning methodology has improved latency by 2.86x (comparing the unpruned B4096 model and two-times pruned B4096 model), while the appropriate choice of DPU (B1600\_lowRAM and B4096) has also contributed a 2.86x improvement.
Together, these factors result in an overall throughput increase of 8.18× from the worst- to best-case scenario, as depicted in Figure \ref{fig:througputRatioBar}.

\subsection{Strategies for Bottleneck Reduction and Performance Optimization}
In the initial design, the application followed a single-stage pipeline: a new cube was generated only after the DPU completed inference on the previous one.
The first column of Table \ref{tab:computationalProfilingStages} shows the time spent on each step, revealing that preprocessing latency is 2.5x higher than DPU latency, making preprocessing the primary bottleneck of the application.

The first strategy to mitigate this bottleneck and improve throughput involved using one POSIX thread (\textit{pthread}) for preprocessing and another for DNN inference.
This parallelization leverages multicore processors, such as the quad-core ARM Cortex-A53, allowing each thread to run on a separate core.
In this setup, synchronization mechanisms are necessary to manage shared resources, prevent race conditions, and ensure data integrity \cite{butenhof1997programming}.

To prevent the preprocessing thread from modifying the cube variable during DPU inference, several schemes were evaluated, including mutexes \cite{butenhof1997programming}, alternating memory buffers, and dedicated buffers for each step.
Considering the significant latency difference between the preprocessing and inference (see first column of Table \ref{tab:computationalProfilingStages}), using separate buffers for each step was chosen to avoid the overhead introduced by mutexes.
Consequently, the cube variable was modified only during the final preprocessing steps, allowing the DPU inference thread to process the previous cube uninterrupted.

\newpage
\begin{table}[h!]
\centering
\caption{Time taken (Averaged over 100 executions) to complete each of the steps of the one/two/three stage whole application execution (Time is in ms).}
\label{tab:computationalProfilingStages}
\resizebox{10cm}{!}{%
\begin{tabular}{cccc}
\toprule
\textbf{Step name / Num. of stages}      & \textbf{1}                               & \textbf{2}              & \textbf{3} \\
\midrule
\textbf{Image loading}                   &  3.255 				 					&   4.790       	  	  &  7.039 \\
\textbf{Cropping and clipping}           &  1.377 				 					&   8.832			  	  &  9.302 \\
\textbf{Reflectance correction}          & 19.337 				 					&  25.789            	  & 37.224 \\
\textbf{Demosaicing}                     &  9.197 				 					&   9.851            	  & 18.164 \\
\textbf{Spatial bilinear interpolation}  & 12.015 				 					&  12.904            	  & 15.409 (87.138) \\ 
\textbf{Cropping + BSQ to BIP}            & 26.548 				 					&  29.930            	  & 52.791 \\
\textbf{Clipping + PN}                   &  5.697 (77.426)		 					&   5.568 (97.664)   	  & 12.309 (65.100) \\
\textbf{DPU Inference*}                  & 32.842$^{\mathrm{a}}$ 					& 61.581$^{\mathrm{b}}$   & 62.217$^{\mathrm{b}}$  \\
\textbf{Longest task time}               & 110.268               					& 97.664                  & 87.138 \\
\bottomrule
\multicolumn{4}{l}{$^{\mathrm{a}}$B4096 DPU, optimum config. $^{\mathrm{b}}$ B3136 DPU, power efficient config.} \\
\multicolumn{4}{l}{$^{\mathrm{*}}$It contains cube transmission times.}
\end{tabular}}
\end{table}

In the two-stage pipeline, preprocessing remained the longest task but was faster than the combined preprocessing and inference time in the single-stage pipeline, leading to improved throughput (see second column of Table \ref{tab:computationalProfilingStages}).
However, running preprocessing and inference in parallel caused each task to slow down individually due to shared resource contention between threads.
The results revealed a load imbalance, indicating an uneven workload distribution across threads that limited overall performance gains.

To further increase throughput, the preprocessing stage was split into two parts: from the start up to spatial bilinear interpolation, and from the final cropping to PN.
his created three concurrent threads: a CPU cube preprocessing thread, a CPU format-adapting thread, and a DPU inference thread.
Synchronization is managed via condition variables, where the preprocessing thread signals the format-adapting thread to start, which then signals the DPU inference thread.
To prevent race conditions, two separate buffers are used for the outputs of both the preprocessing and format-adapting threads.

Since the first two stages run on the ARM processor, they share the 32 128-bit SIMD/floating-point registers.
Introducing an additional thread reduces the available resources per thread, causing the total latency of these stages to be higher than in the single-stage pipeline.
However, throughput improves because previously sequential operations now run in parallel.
Results shown in the third column of Table \ref{tab:computationalProfilingStages} demonstrate a 15\% (17 ms) reduction in the latency of the longest task.

\subsection{Comprehensive Selection of Deep Processing Unit}
Regardless of whether one, two, or three stages are used, preprocessing remains the system bottleneck, limiting overall throughput.
Therefore, selecting the DPU configuration with the lowest latency is not always optimal, as it demands more hardware resources and higher power consumption.
Moreover, because the input image size is 192x384x25 bytes (1.8 MB) and skip connections increase the size of deeper layer feature maps, not all activation maps fit in on-chip BRAM.
Increasing the number of memory banks for activations from 16 (2 MB) to 24 (3 MB) yields a 4 ms performance improvement (see Table \ref{tab:profilingDPUversion}, columns two and three).

It is important to note that the B2304 configuration does not meet the inference latency requirements needed to prevent the DPU from becoming the bottleneck when data synchronization tasks are included.
As a result, the final DPU configuration uses a B3136 core with low RAM usage, achieving 10.54 FPS for the entire application pipeline.
As shown in Table \ref{tab:resourcesDPUversion}, this configuration reduces resource usage by 20.28\% fewer DSPs, 16.67\% fewer URAMs, 3.38\% fewer registers, and 16.69\% fewer flip-flops.

\newpage
\begin{table}[h!]
\centering
\caption{Computational profiling (Averaged over 100 executions) of three stage whole application with the DPU configurations of Table \ref{tab:resourcesDPUversion} (Time is in ms).}
\label{tab:profilingDPUversion}
\resizebox{8cm}{!}{%
\begin{tabular}{ccccc}
\toprule
\textbf{Parameter / DPU name}   & \textbf{B4096} & \textbf{B3136} & \textbf{B3136} & \textbf{B2304} \\
\midrule
\textbf{Cube preprocessing}     &   89.786 & 88.931  &  87.138 &  85.554 \\
\textbf{Format adapting}        &   67.695 & 66.261  &  65.100 &  65.740 \\
\textbf{DPU inference}          &   19.241 & 38.865  &  42.959 &  54.897 \\
\textbf{Avg. bandwidth (MB/s)}  & 1759.738 & 882.568 & 798.328 & 625.645 \\
\bottomrule
\end{tabular}}
\end{table}

Power consumption on the K26 SOM was monitored using the \textit{platformstats} application, which interfaces with the INA260 current sensor via I2C.
The measured power consumption was 2.44W when the PL is powered down in idle state, 3.2W when the PL was powered up but not programmed, and 3.52W when the PL was powered up and programmed but not running.
During application execution, average power consumption was 5.2W, almost 2W lower than previous results.

\section{Conclusions}\label{sec:conclusions}
While the use of DNNs for HSI has been intensively investigated in recent years, their application to ADS and autonomous navigation remains particularly challenging.
Integrating DNNs with the rich spectral information from snapshot HSI cameras significantly enhances intelligent vision capabilities by addressing metamerism issues inherent in traditional RGB systems.
This paper addresses the main challenges in developing embedded processing architectures suited for deploying HSI-based intelligent vision systems in ADS.
It is demonstrated that acquiring richer and more accurate spectral data with modern snapshot hyperspectral sensors enables the use of aggressive compression techniques without sacrificing performance.

Snapshot filter-on-chip HSI technology, capable of capturing hyperspectral data at video rates, requires a sophisticated processing pipeline to convert raw 2D images into 3D data cubes compatible with DNN input.
Although often overlooked, this preprocessing stage is a critical bottleneck, as confirmed in this design.
To address this, a refined hardware/software co-design approach is proposed that balances task distribution, identifies inefficiencies, and enhances the overall performance on FPGA-based SoC platforms.

For cube preprocessing, computational efficiency was maximized by combining thread-level and data-level parallelism on the quad-core ARM Cortex-A53 processor.
The memory representation analysis considers constraints such as the 2D input format, DPU requirements, and the order and nature of preprocessing operations, which favour BSQ format initially and BIP format later for DPU compatibility.
This approach avoids performance penalties from premature format conversion.

Regarding compression techniques, an iterative pruning method based on static and dynamic analysis demonstrated a two-order-of-magnitude reduction in parameters and a one-order-of-magnitude decrease in OPS, achieving results comparable to SotA models in RGB benchmarks but with significantly lower computational complexity.
A layer-by-layer comparison between pre-training and post-training pruning highlights opportunities to combine insights from each stage to develop more effective pruning techniques for segmentation, an area that remains under active development.
The combined impact of pruning and DPU architecture selection was evaluated in terms of power consumption (W), logical resource usage (DSP units, LUTs), memory footprint (MB), model complexity (OPS), and throughput (FPS).

To optimize the system pipeline, preprocessing was identified as the primary bottleneck.
The system was therefore restructured into three concurrently executing stages: two preprocessing stages and one inference stage.
This reorganization, together with pruning-based complexity reduction, enabled the selection of a more efficient DPU architecture without compromising model performance.

In conclusion, this work demonstrates that applying a hardware/software co-design approach and targeted optimization techniques enables the deployment of an HSI-based embedded processing architecture that enhances the robustness of intelligent vision systems for ADS.
Nevertheless, further acceleration of raw image preprocessing is required, either through specialized hardware or by integrating this stage into the DNN feature extraction module.
Additionally, obtaining low-complexity compressed models will facilitate exploration of stream or dataflow-type accelerators, where each neural network layer is mapped as a distinct hardware unit and interconnected in a stream-like manner to increase inference throughput.

\section{Acknowledgments}
This work was partially supported by the University of the Basque Country (UPV/EHU) under grant GIU21/007, by the Basque Government under grant PRE\_2024\_2\_0154 and by Union Europea-NextGenerationEU through the Cátedras Chip program, SOC4SENSING TSI-069100-2023-0004.

\bibliographystyle{unsrtnat}
\bibliography{references}

\end{document}